\pdfoutput=1
\documentclass[11pt]{article}
\usepackage[final]{acl}
\usepackage{times}
\usepackage{multirow}
\usepackage{latexsym}

\usepackage[T1]{fontenc}
\usepackage[utf8]{inputenc}
\usepackage{amsmath, amssymb, amsfonts}
\usepackage[ruled,linesnumbered]{algorithm2e}
\SetKwComment{Comment}{/* }{ */}
\SetKwFunction{Range}{range}
\usepackage{microtype}
\usepackage{subcaption}
\usepackage{inconsolata}

\usepackage{graphicx}
\usepackage{tikz-qtree}
\usepackage{tikz-dependency}
\newcommand*{\Nearrow}{\rotatebox[origin=c]{45}{\(\Rightarrow\)}}
\newcommand*{\Nwarrow}{\rotatebox[origin=c]{-45}{\(\Leftarrow\)}}
\usepackage[usenames,dvipsnames]{xcolor}
\usepackage{fontawesome}
\usepackage{bm}
\newcommand*\bell{\pmb{\ell}}

\title{Bringing Emerging Architectures to Sequence Labeling in NLP}
\author
{
    Ana Ezquerro\SP{1,2} \\ \texttt{ana.ezquerro} \And Carlos Gómez-Rodríguez\SP{1} \\ \texttt{carlos.gomez} \And David Vilares\SP{1} \\ \texttt{david.vilares} \AND \\[-1cm]
    \SP{1}Universidade da Coru\~{n}a, CITIC  (\texttt{@udc.es})\\
    $^2$Graz University of Technology, IML (\texttt{@tugraz.at})\\
}

\def\SPSB#1#2{\rlap{\textsuperscript{#1}}\SB{#2}}
\def\SP#1{\textsuperscript{#1}}
\def\SB#1{\textsubscript{#1}}

\usepackage{fontawesome}
\newcommand{\xlstm}{{\faFlash}}
\newcommand{\baseline}[1]{\textcolor{gray}{#1}}

\usepackage{tikz}
\usetikzlibrary{arrows.meta}
\newcommand{\DiagonalArrow}{%
    \begin{tikzpicture}[baseline=0pt] 
        \draw [-{Latex[open]}] (0,0) -- (0.25,0.2); 
    \end{tikzpicture}%
}

\begin{document}
\maketitle
\begin{abstract}
Pretrained Transformer encoders are the dominant approach to sequence labeling. While some alternative architectures---such as xLSTMs, structured state-space models, diffusion models, and adversarial learning---have shown promise in language modeling, few have been applied to sequence labeling, and mostly on flat or simplified tasks. We study how these architectures adapt across tagging tasks that vary in structural complexity, label space, and token dependencies, with evaluation spanning multiple languages. We find that the strong performance previously observed in simpler settings does not always generalize well across languages or datasets, nor does it extend to more complex structured tasks.
\end{abstract}

\section{Introduction}

Sequence labeling (SL) is a problem in machine learning, and particularly in NLP, where each element in a sequence is assigned exactly one output label. A feature of SL tasks is that labels are not predicted in isolation: they often depend on neighboring inputs. To address these dependencies, various architectures have been proposed over the years to model both short- and long-range interactions between input tokens, including Conditional Random Fields \citep[CRF;][]{lafferty2001conditional}, sliding-window perceptrons \cite{zhang-clark-2008-joint}, Long Short-Term Memory networks \citep[LSTMs;][]{hochreiter1997long}, and Convolutional Neural Networks \citep[CNNs;][]{lecun1995convolutional}. Today, Transformers~\cite{NIPS2017_3f5ee243}, pretrained on large data, are the dominant approach for tagging tasks, offering superior ability to model relations across words.

At the same time, research continues to explore both alternatives to Transformers and architectural variants of them across a range of machine learning tasks. This includes, for instance, enhanced contextualization mechanisms \citep[xLSTM;][]{beck-etal-2024-xlstm} and structured state-space models \citep[Mamba;][]{gu-etal-2022-s4} for language modeling, as well as diffusion models \citep{ho2020denoising} and generative adversarial networks \citep[GANs;][]{goodfellow2014generative}, which were originally developed for computer vision applications. Some of these architectures have shown competitive, albeit preliminary, performance compared to Transformers in metrics such as perplexity (see again \citeauthor{beck-etal-2024-xlstm}) and could be promising for tagging tasks. Meanwhile, others have already been adapted for sequence labeling in relatively simpler settings \citep[e.g.,][]{huang-etal-2023-diffusionsl,tong-etal-2024-generative}, typically focusing on tasks like named-entity recognition (NER), part-of-speech (PoS) tagging, or (Chinese/Japanese/Korean) word segmentation. However, evaluations have mostly remained within these flatter, coarse-grained setups, leaving open questions about how well these models generalize to more fine-grained or complex scenarios, such as recent linearizations for structured tasks involving trees \citep{gomez-rodriguez-vilares-2018-constituent, kitaev-klein-2020-tetra, amini-etal-2023-hexatagging} or graphs \citep{ezquerro-etal-2024-dependency}.

\paragraph{Contribution} 
We investigate alternative architectures beyond pretrained Transformers for sequence labeling, focusing on architectures not originally designed for token-level classification but with potential for this task. Specifically, we explore how these models can be adapted to capture linguistic structure in sequence-labeling problems, considering tasks of varying complexity, output label spaces, and dependency spans. Rather than aiming to outperform Transformers universally, the goal is to better understand the relative strengths and limitations of these architectures across different problem types. Our results show that bidirectional xLSTM architectures are generally superior to the traditional BiLSTMs across tasks and datasets, although they still fall behind Transformers. Diffusion tagging and state-space models trail the baseline, suggesting they may be less suited for NLP tagging. Finally, adversarial tagging yields a noteworthy result, consistently rivaling or surpassing the Transformer baseline across diverse tasks, including the most complex structured settings. Our code is publicly available at \url{https://github.com/anaezquerro/separ}.

\section{Background}
We now outline concepts in tagging and modeling advances.

\subsection{Sequence labeling for NLP}\label{sect:rw-sequence-labeling}
Classical problems like PoS tagging or lemmatization naturally align with the definition of sequence labeling. Others, such as NER, chunking \citep{ramshaw-marcus-1995-text}, segmentation \citep{hacioglu-etal-2004-semantic}, semantic role labeling \cite{strubell-etal-2018-linguistically} and slot filling \citep{li-etal-2020-handling}, can also be framed as tagging problems, typically using lightweight encoding schemes that assign token-level labels (e.g. IOB encoding). Despite gains from pretrained Transformers, many tasks, especially simpler ones, were already tractable with shallow, easier-to-deploy models.

Similarly, previous efforts focused on reformulating tree- and graph-structured tasks as tagging through linearizations. However, pre-neural models struggled with these problems. \citet{spoustova2010dependency} showed that linguistically informed linearizations trained with pre-neural models performed impractically compared to the state of the art at the time. This limitation of earlier models started to change with context-aware encoders based on BiLSTMs or Transformers, which have recently revived sequence labeling for structured prediction, though effectively modeling such structure was not immediate \cite{li-etal-2018-seq2seq}. In addition, they open up new possibilities for evaluating alternative sequence tagging architectures under more demanding testbeds, with large output spaces and long-range dependencies. In this context, linearization strategies have been proposed for both continuous \cite{gomez-rodriguez-vilares-2018-constituent, kitaev-klein-2020-tetra, amini-cotterell-2022-parsing} and discontinuous constituent parsing \cite{vilares-gomez-rodriguez-2020-discontinuous}, as well as for projective and non-projective syntactic dependency parsing \cite{strzyz-etal-2019-viable, amini-etal-2023-hexatagging} and, recently, graph parsing~\cite{ezquerro-etal-2024-dependency}.

\subsection{Sequence modeling}\label{sect:rw-modeling}

While LSTMs and Transformers are the standard encoders for token-level classification tasks, recent years have seen growing interest in alternative techniques---some of which still leverage self-attention---and architectures that replace the Transformer with different contextualization systems. Although many of these methods were not initially designed for sequence labeling, we now examine their potential. 

GANs, created for image generation, have been extended to text generation in NLP, addressing the challenge posed by the discrete nature of language \cite{kusner-etal-2016-gan,zhang-etal-2017-seqgan}. However, fewer studies have explored their use in non-generative tasks, which require generating or refining structured outputs rather than free-form text. Notably, \citet{parnow-etal-2021-grammatical} trained a GAN-based tagging system to enhance grammatical error correction by enriching the learning process with generated errors. \citet{tong-etal-2024-generative} improved word segmentation and NER with a GAN-based framework, using the generator as a labeler and a discriminator to guide accurate sequences.

Similarly, diffusion models \cite{ho2020denoising} are often used for generative NLP tasks \cite{he-etal-2023-diffusionbert, han-etal-2023-ssd}. While several studies improve denoising embeddings in latent space \cite{gao-etal-2024-empowering, zhou-etal-2024-diffusion} to enhance text-to-text generation \cite{shi-etal-2023-diffusion, liu-etal-2024-p3sum}, few have explored their potential for tagging. Notably, some recent work has adapted diffusion processes for NER \cite{shen-etal-2023-diffusionner} and PoS tagging \cite{huang-etal-2023-diffusionsl}.

In addition, recent work has explored alternative architectures replacing self-attention. \citet{gu-etal-2022-s4} introduced structured-state space models (SSM) for linear-time language modeling, addressing the quadratic complexity of Transformers. \citet{beck-etal-2024-xlstm} introduced the xLSTM, a variant of the LSTM with parallelizable capabilities and better modeling of dependencies. Still, xLSTM has been mainly evaluated on language modeling and bioinformatics \cite{heidari2025study,sun2025molgraphxlstmgraphbasedduallevelxlstm}, not on tagging for NLP.

\section{Sequence labeling architectures}

Current SL models typically consist of two main components. First, an encoder $\mathcal{E}_\theta: \mathcal{V}^n \to \mathbb{R}^{n\times d}$, usually a pretrained masked language model, contextualizes an input sentence $W=(w_1\cdots w_n) \in \mathcal{V}^n$ in a $d$-dimensional latent space. Then, each token embedding is passed through a decoder\footnote{E.g., either a feed-forward network (FFN) with a non-linear activation or more complex projection heads (e.g., CRFs) could be used.} $\mathcal{D}_\phi: \mathbb{R}^{n\times d} \to \mathbb{R}^{n\times |\mathcal{L}|}$ to produce a probability distribution over the label set $\mathcal{L}$.

This work examines alternative formulations of tagging along two complementary directions: modifying the learning strategy (typically a supervised input-to-label mapping) and the main underlying architecture for encoding (typically a pretrained language model). We begin by discussing strategies for modeling the label space $\mathcal{L}$, focusing on stable diffusion (\S\ref{subsec:diffusion}) and adversarial learning (\S\ref{subsec:adversarial}). Although these techniques have been applied to tagging \cite{huang-etal-2023-diffusionsl, tong-etal-2024-generative}, prior evaluations were often limited in scope, typically restricted to simpler tasks. We then explore SSM models and xLSTMs as alternatives to Transformers for sequence contextualization in tagging problems (\S\ref{subsec:enhanced}).

\subsection{Diffusion Tagging}\label{subsec:diffusion}
We first present a sequence labeler using diffusion tagging, based on a bit-tag converter \cite{huang-etal-2023-diffusionsl} to handle discrete sequential outputs. \citet{huang-etal-2023-diffusionsl} used the 
denoising diffusion implicit model (DDIM)
sampling\footnote{\citet{song-etal-2021-denoising} proposed a faster, more stable reformulation of the original denoising process \cite[DDPM]{ho2020denoising} using non-Markovian deterministic sampling.} to directly predict the target data, thus deviating from the original step-by-step denoising process of diffusion models. To better align the principles of stable diffusion to neural tagging,  we adopt the bit-tag converter of \citet{huang-etal-2023-diffusionsl}  but propose a  conditional diffusion model that iteratively denoises a random signal by learning the added noise during the forward process, closely following the original denoising process for a fuller evaluation of diffusion in tagging.

\paragraph{Bit conversion} The Bit-Tag converter (BT) by \citet{huang-etal-2023-diffusionsl} transforms an input tag sequence $(\ell_1\cdots\ell_n)\in\mathcal{L}^n$ into a sequence of bits to treat the output as a continuous signal. Formally, the forward transformation (\texttt{tag2bit}) maps each integer identifier of a discrete set $\mathcal{L}$ into a sequence of $m=\lceil \log_2 |\mathcal{L}|\rceil$ bits. For instance, given $\mathcal{L}_4=\{0,1,2,3\}$, the \texttt{tag2bit} operation transforms each label into a sequence of 2 bits, so $\texttt{tag2bit}(\mathcal{L}_4)=\{00, 01, 10, 11\}$. The reverse process (\texttt{bit2tag}) transforms a sequence of $m$ bits into an integer in the range $[0,2^m-1]$.\footnote{If $\mathcal{L}$ is not power of 2, \texttt{bit2tag} may yield integers outside $\mathcal{L}$. These are removed and replaced by the most common tag.}

\paragraph{Forward process} 
Diffusion models gradually add Gaussian noise to a clean sample $\mathbf{x}_0$ during $T$ timesteps, and train a neural network to model the reverse process, progressively denoising the sample. When adapting diffusion models to discriminative tasks the input to the forward process is the actual target, and the input sentence is fed as a conditional signal. In this case, $\mathbf{x}_0$  is the noise-free bit representation of the target sequence of labels. Then, from a noise schedule $\beta_1\cdots\beta_T$, where $\beta_i < \beta_{i+1}$ and $\beta_i \in (0,1)$, $\forall i=1\cdots T$, the sequence of latent variables $\mathbf{x}_1\cdots \mathbf{x}_T$ follows a Markov process, such that each latent variable is generated by adding Gaussian noise to the previous one
(Equation \ref{eq:diffusion-beta}).  When defining $\alpha_t=1-\beta_t$ and $\bar{\alpha}_t=\prod_{s=1}^t\alpha_t$, the forward process is defined conditioned on $\mathbf{x}_0$ (Equation \ref{eq:diffusion-alpha}).
\begin{equation}\label{eq:diffusion-beta}
    q(\mathbf{x}_t|\mathbf{x}_{t-1}) \sim \mathcal{N}(\sqrt{1-\beta_t}\mathbf{x}_{t-1}, \beta_t\mathbf{I})
\end{equation}
\begin{equation}\label{eq:diffusion-alpha}\small
    q(\mathbf{x}_t|\mathbf{x}_0) \sim \mathcal{N}(\sqrt{\bar{\alpha}_t}\mathbf{x}_0, (1-\bar{\alpha}_t)\mathbf{I})
\end{equation}

The diffusion tagger (DiT) trains a neural network to estimate the noise component present in the latent variable at each timestep, using: (i) the latent variable $\mathbf{x}_t$, (ii) the timestep $t$ and (iii) the input tokens $(w_1\cdots w_n)$ as a conditional signal. During training, timesteps are uniformly sampled to generate latent variables following Equation \ref{eq:diffusion-alpha}. Following \citet{ho2020denoising}, our model is tasked to minimize the MSE loss between the real and predicted noise.

Algorithm \ref{alg:diffusion-forward-pass} and Figure \ref{fig:diffusion-training} (\S \ref{ap:figures}) show the forward process for a sample $(w,\ell)$. The token encoder $\mathcal{E}_\theta$ is an encoder-only network that contextualizes tokens in the input sequence. The decoder $\mathcal{D}_\phi$ is a Transformer-based architecture that accepts as input the (noised) sample, the timestep $t$ and the contextualized embeddings. Following the bit conversion described above, each tag is represented as an $m$-dimensional binary vector. At each training step, a timestep $t$ is sampled uniformly to compute the latent variable $\mathbf{x}_t$. The decoder is then trained to estimate the noise added to $\mathbf{x}_t$.

\paragraph{Denoising process} 
For our diffusion tagger we adopt DDIM sampling \cite{song-etal-2021-denoising}, which allows skipping timesteps to increase inference speed. Since each latent variable is defined in the forward process as $\mathbf{x}_t=\sqrt{\bar{\alpha}_t}\mathbf{x}_0 + \sqrt{1-\bar{\alpha}_t}\mathbf{e}$, the estimation of a previous latent $\mathbf{x}_{k}$, where $k<t$, can be obtained with the estimated noise from $\mathcal{D}_\phi$. Algorithm \ref{alg:diffusion-denoise} and Figure \ref{fig:diffusion-inference} (\S\ref{ap:figures}) show the denoising process using an hyperparameter $s$, controlling how many timesteps are skipped in the reverse process.

\begin{algorithm}[tpb!]
\begin{small}
\caption{Forward process.}\label{alg:diffusion-forward-pass}
Noise schedule $\{\bar{\alpha}_1\cdots\bar{\alpha}_T\}$\;
Word encoder $\mathcal{E}_\theta:\mathcal{V}^n\to\mathbb{R}^{n\times d}$\;
Decoder $\mathcal{D}_\phi:(\mathbb{R}^{n\times m}, \mathbb{N}, \mathbb{R}^{n\times d})\to \mathbb{R}^{n\times m}$\;
\ForEach{$(w,\ell)$ in dataset}{
$\mathbf{w}=\mathcal{E}_\theta(w)$  {\small\Comment*[r]{word embeddings}}
Initial signal:
$\mathbf{x}_0=\texttt{tag2bit}^*(\ell)$\;
$t\sim \mathrm{Unif}(1,T)$  {\small\Comment*[r]{sample timestep}}
$\mathbf{e}\sim \mathcal{N}(\mathbf{0}, \mathbf{I})$  {\small\Comment*[r]{Gaussian noise}}
Latent variable: $\mathbf{x}_t=\sqrt{\bar{\alpha}}_t \mathbf{x}_0+\sqrt{1-\bar
{\alpha}_t}\mathbf{e}$\;
Gradient descent step on: $\nabla_{\theta,\phi}=\|\mathbf{e}- \mathcal{D}_\phi (\mathbf{x}_t,t,\mathbf{w})\|^2$
}
\end{small}
\end{algorithm}
\begin{algorithm}[tpb!]
\begin{small}
\caption{Denoising process.}\label{alg:diffusion-denoise}
\KwIn{Sample $W=(w_1\cdots w_n)\in\mathcal{V}^n$ and number of skipped inference steps $s$.}
\KwOut{Estimated tag sequence $\tilde{\ell}$.}
$\mathbf{w}=\mathcal{E}_\theta(w)$  {\small\Comment*[r]{word embeddings}}
$\mathbf{x}_T\sim \mathcal{N(\mathbf{0}, \mathbf{I})}$ {\small\Comment*[r]{Gaussian noise}}
$t\leftarrow T$\;
\While{$t>0$}{
    $\tilde{\mathbf{e}}_t  = \mathcal{D}_\phi(\mathbf{x}_t,t,\mathbf{w})$\;
    $k=t-s \text{ if } t-s > 0 \text{ otherwise } 0 $\;
    $\tilde{\mathbf{x}}_{k} = \frac{\sqrt{\bar{\alpha}_{k}}}{\sqrt{\bar{\alpha}}_t} (\mathbf{x}_t-\sqrt{1-\bar{\alpha}_t}\tilde{\mathbf{e}}_t )+\sqrt{1-\bar{\alpha}_{k}}\tilde{\mathbf{e}}_t$\;
    $t \leftarrow k$
}
$\tilde{\ell} = \texttt{bit2tag}^*(\tilde{\mathbf{x}}_0)$ {\small\Comment*[r]{bit conversion}}
\KwRet{$\tilde{\ell}$}
\end{small}
\end{algorithm}

\paragraph{Model architecture} We use a pretrained  masked language model for the encoder module $\mathcal{E}_\theta$, recovering the last hidden states as token embeddings. The decoder $\mathcal{D}_\phi$ learns the added noise from a latent variable $\mathbf{x}_t$, the token embeddings and the timestep $t$, which represents the noise level applied to $\mathbf{x}_t$. We adopt the DiT block proposed by \citet{huang-etal-2023-diffusionsl}, which relies on a learnable embedding layer $\tau$ to represent the timestep $t$. The latent variable $\mathbf{x}_t$, the word embeddings $\mathbf{w}$ and the time embedding $\tau(t)$ are merged and fed to  a stack of Transformer layers with residual connections. The final layer has an output dimension of $m$ with a linear activation to predict the added noise.

\subsection{Adversarial Tagging}\label{subsec:adversarial}

Next, we follow the approach by \citet{tong-etal-2024-generative} to build an adversarial tagger composed of two modules: a generator $G_\psi$ and a discriminator $D_\varphi$. In their concept of adversarial training for tagging, the generator receives a sentence and is trained to generate the tag sequence, while the discriminator evaluates the generator's predictions against the ground-truth tags to identify incorrect outputs. To simplify the original setup and enable clearer comparison with other taggers, we remove the CRF module and reduce the architecture of the discriminator to a 2-layered BiLSTM stack.

\paragraph{Generator} The generator $G_\psi:\mathcal{V}^n \to \mathbb{R}^{n\times |\mathcal{L}|}$ is an encoder-decoder neural architecture that learns the real tags from an input sentence, as traditional SL approaches. The encoder $G_\psi^\mathcal{E}:\mathcal{V}^n\to\mathbb{R}^{n\times d}$ maps each token into a learned latent space, and the decoder $G_\psi^\mathcal{D}:\mathbb{R}^d\to\mathbb{R}^{|\mathcal{L}|}$ independently projects each embedding into the learned tag distribution. The generator loss is defined as the cross-entropy between the real tags and the predicted distribution.

\paragraph{Discriminator} The discriminator $D_\varphi:(\mathcal{V}^n, \mathbb{R}^{n\times |\mathcal{L}|})\to \mathbb{R}^{n}$ takes as input the sequence of words and an estimated distribution over $\mathcal{L}$; and outputs a similarity score measuring how close the predicted distribution is to the true distribution $p(\ell|w)$.  Let $G_\psi(w) =(\tilde{\bell}_1,...,\tilde{\bell}_n)$ be the predicted distribution of the generator and $\mathbf{L}=(\bell_1,...,\bell_n)=\texttt{onehot}^*(\ell_1,...,\ell_n)$ the one-hot representation of the real tag sequence.  To ease backpropagation, we apply the Gumbel-Softmax relaxation \cite{jang-etal-2017-categorical} to smooth the one-hot representation of the target tags, following \citet{tong-etal-2024-generative}. The discriminator loss models valid tag sequences conditioned on the input to spot incorrect tags. Intuitively, it approximates $D_\varphi(w,\mathbf{L})$ to $\mathbf{1}$, and  $D_\varphi(w,G_\psi(w))$ to $\mathbf{s}=(s_1,..,s_n)$, where each value is defined as in Equation \ref{eq:discriminator-score}:

\begin{equation}\label{eq:discriminator-score}\small
    s_i=\begin{cases}
        1 & \text{if }\arg\max\{\tilde{\bell}_i\} = \ell_i\\
        0 & \text{otherwise}
    \end{cases}
\end{equation}
using $\mathcal{H}$ as the cross-entropy loss (Equation \ref{eq:discriminator-loss}):
\begin{equation}\label{eq:discriminator-loss}\small
\begin{aligned}
    L_{D,p} &= \mathcal{H}\big(D_\varphi(w,\mathbf{L}),\mathbf{1}\big) \\
    L_{D,G} &= \mathcal{H}\big( D_\varphi(w, G_\psi(w)), \mathbf{s} \big)
\end{aligned}
\end{equation}

\paragraph{Adversarial training} 
To replicate the adversarial dynamics of GANs, we compute the adversarial loss (Equation \ref{eq:adversarial-loss}) to guide $G_\psi$ to generate distributions that challenge $D_\varphi$. The generator loss \( L_G \) (Equation~\ref{eq:adversarial-training}) includes an hyperparameter \( \lambda \) that controls the influence of the adversarial component during training. By adjusting $\lambda$, the generator is encouraged not only to match the gold tags but also to produce outputs that fool the discriminator. Figure~\ref{fig:adversarial} (\S\ref{ap:figures}) visualizes our adversarial training.

\begin{equation}\label{eq:adversarial-loss}\small
    L_A = \mathcal{H}\big( D_\varphi(w, G_\psi(w)), \mathbf{1} \big)
\end{equation}
\begin{equation}\label{eq:adversarial-training}\small
    \begin{aligned}
      L_G &=\mathcal{H}(G_\psi (w), \ell) + \lambda L_A \\
     L_D &= L_{D,p} + L_{D,G}
    \end{aligned}
\end{equation}

\paragraph{Model architecture} 
For the generator module, we rely on a encoder-decoder architecture, where the decoder is a FFN with a non-linear activation. For the discriminator, we use a lightweight BiLSTM-based encoder\footnote{Initial experiments showed no significant performance drop when reducing discriminator size, matching results without doubling the network as in \citet{tong-etal-2024-generative}.} to contextualize the sequence of token embeddings and tag logits, followed by a FFN to validate the correctness of the input tag distribution.

\subsection{Alternatives for sequence modeling}\label{subsec:enhanced}
We now describe encoder architectures beyond Transformers that, to our knowledge, remain untested for tagging tasks.

\paragraph{xLSTM} \citet{beck-etal-2024-xlstm} recently proposed a new recurrent unit inspired on the LSTM unit that deals with the main drawbacks of its ancestor when modeling long dependencies and enabling parallelization. The xLSTM relies on two blocks: the sLSTM, which deals with the first problem by stabilizing the information flow in the forget gate; and the mLSTM, which enables GPU parallelization by replacing the vectorial form of the hidden states with learnable matrices (resembling the self-attention operation). By stacking xLSTM blocks, we explore xLSTM-based encoders for contextualizing input sentences for tagging tasks. Additionally, inspired on the BiLSTM design \cite{graves-etal-2005-bilstm}, we explore the \textbf{BixLSTM}, which processes with two different xLSTM units an input sequence from left-to-right and right-to-left and concatenates their representations.

\paragraph{\textsc{Mamba-2} (SSD)} Building on the structured state-space (S3) framework, \citet{gu-etal-2022-s4} introduce the structured state-space sequence model (S4), which captures long-range dependencies with linear time and space complexity by parameterizing the dynamics using diagonal state matrices and computing convolution kernels in the frequency domain, thus enabling efficiency and expressiveness to sequence modeling. More recently, \citet{gu-etal-2023-mamba} proposed \textsc{Mamba}, a selective state-space model (S6) extended from S4  with dynamic input-dependent weights and gating mechanisms. As part of our comparison, we adopt \textsc{Mamba-2} \cite{dao-etal-2024-transformers},  an improved version of S6 with architectural refinements that address the numerical instability and throughput limitations of earlier SSMs, in the encoder architecture to examine how its state-space dual framework (SSD) behaves in relation to Transformer-based models.

\section{Experiments}\label{sec:experiments}

We evaluate these architectures on a multilingual benchmark covering multiple  tasks, opting for datasets that cover scenarios of varying complexity, different output vocabulary sizes  and a range of token dependencies, as these factors may influence how the underlying architecture affects performance. See \S\ref{ap:treebanks} (Table~\ref{tab:treebanks}) for details on the datasets used, and \S\ref{ap:examples} for examples of the parsing linearizations introduced in next paragraphs. We have aimed to maintain a relatively homogeneous set of languages across tasks, but this was not always possible due to dataset availability or evaluation setup. 

\paragraph{PoS tagging} We use datasets with coarse- and fine-grained tag sets, as well as morphologically rich tags for languages with complex morphology: eight Universal Dependencies (UD) treebanks \cite{nivre-etal-2020-universal}, the Penn Treebank (PTB; \citealp{marcus-etal-1993-building}), the Chinese Treebank (CTB; \citealp{xue-etal-2005-chinese}), and the Statistical Parsing of Morphologically Rich Languages datasets (SPMRL; \citealp{seddah-etal-2014-introducing}).\footnote{For SPMRL, we predict PoS tags as a single task to maintain a consistent setup across all experiments. While these tags are typically decomposable into independent morphological components suitable for multi-task learning, our simplified approach explains why the reported performance is lower than standard benchmarks.}

\paragraph{NER} We evaluate the following datasets
spanning diverse languages using the BIO scheme annotation: CoNLL-2003 \cite{tjong-kim-sang-de-meulder-2003-introduction}, BioNLP \cite{pyysalo-etal-2013-overview}, DrugsNER \cite{drugs-dataset}, EIEC \cite{alegria-etal-2004-design}, GermEval-2014 \cite{benikova-etal-2014-germ}, HiNER \cite{murthy-etal-2022-hiner}, JBNLPA \cite{collier-kim-2004-introduction}, KLUE \cite{sungjoon-etal-2021-klue}, NYTK-NerKor \cite{eszter-vadasz-2021-nytk}, Webbnyheter \cite{text-etal-2024-ner}, Weibo-NER \cite{peng-dredze-2015-named} and WikiNER \cite{nothman-etal-2012-wikiner}.

\paragraph{Constituent parsing} We use the PTB \cite{marcus-etal-1993-building}, the CTB \cite{xue-etal-2005-chinese}, and the SPMRL datasets \cite{seddah-etal-2014-introducing}. To cast constituent parsing as tagging, we adopt two linearizations: the relative encoding~\citep[\textbf{R},][]{gomez-rodriguez-vilares-2018-constituent}, which captures the difference in common ancestors between words, and  the tetra-tagging~\citep[\textbf{T},][]{kitaev-klein-2020-tetra}, based on child direction in a binary tree. These represent the two families of constituent linearizations: depth- and transition-based.

\paragraph{Dependency parsing} 
We use the same eight UD treebanks \cite{nivre-etal-2020-universal} as for PoS tagging. We also evaluate the dependency taggers on the PTB and CTB using the dependency conversion proposed by \citet{de-marneffe-manning-2008-stanford}. Among existing linearizations to cast dependency parsing as a tagging task, the absolute encoding~\citep[\textbf{A},][]{strzyz-etal-2019-viable} is the simplest, as each label independently encodes the head of a word. Bracketing encodings represent trees using balanced bracket strings distributed across labels. In the case of naive bracketing~\citep[\textbf{B},][]{strzyz-etal-2019-viable}, the label set is unbounded; we also consider the bounded 4-bit and 7-bit variants (\textbf{B}\textsubscript{4}, \textbf{B}\textsubscript{7}), which reduce the label space~\citep{gomez-rodriguez-etal-2023-4}. Finally, hexatagging~\citep[\textbf{H},][]{amini-etal-2023-hexatagging} encodes a constituency-based transformation of the dependency tree.

\paragraph{Graph parsing} 
While linearizations for dependency trees are well studied, \citet{ezquerro-etal-2024-dependency} recently proposed unbounded and bounded variants for dependency graphs, a more expressive formalism allowing reentrancies, cycles, and non-connectivity. With $k$ being the number of planes (see \S\ref{appendix-graph-parsing}), we use the relative (\textbf{R}), bracketing with $k=3$ (\textbf{B}), $4k$-bit with $k=4$ ($\boldsymbol{4k}$) and $6k$-bit ($\boldsymbol{6k}$) encodings and conducted experiments in multiple languages from the SDP \citep{oepen-etal-2015-semeval} and IWPT \citep{bouma-etal-2021-raw} corpus.

\paragraph{Evaluation} We use accuracy for PoS tagging; mention-level F1 score for NER; UAS and LAS for dependency parsing; labeled F1 with the \texttt{EVALB}\footnote{\href{https://nlp.cs.nyu.edu/evalb/}{nlp.cs.nyu.edu/evalb}.} tool and the \texttt{COLLINS.prm} and \texttt{SPMRL.prm} parameter file for constituency parsing; and labeled F1 with the \texttt{sdp-toolkit}\footnote{\href{https://github.com/semantic-dependency-parsing/toolkit}{github.com/semantic-dependency-parsing/toolkit}.} for graph parsing.

\paragraph{Model configuration} Our diffusion and adversarial taggers finetune XLM-RoBERTa\textsuperscript{L} \citep[XLM;][]{liu-etal-2020-roberta} to produce word embeddings. The diffusion tagger uses XLM as the encoder and stacks 6 DiT blocks with 16 heads as the decoder, where the bit sequence is introduced as target and the word and time embeddings as conditional signal. We set $T=100$ and $s=10$ steps for inference and a linear variance schedule with $\beta_1=0.002$ and $\beta_T=0.03$, in order to adjust the noise level to the bit range in larger timesteps. The adversarial tagger finetunes a XLM encoder for the generator, stacked with a FFN to output the tag distribution, and a 2-layered BiLSTM stack for the discriminator, stacked with a FFN to predict the correctness of the two input sequences. The hyperparameter $\lambda$ is fixed to 1. For the xLSTM encoder, we stacked four xLSTM blocks of hidden size $d=400$, where each block consists of an mLSTM followed by an sLSTM. To allow bidirectionality, the BixLSTM block processes the input sequence in both directions using two xLSTMs (each $d$=200), and outputs the concatenation of their representations to the next layer. For the SSD-based encoder, we finetuned \textsc{Mamba2-370m} \cite{dao-etal-2024-transformers} using a FFN to predict the final tag sequence.


\paragraph{Baseline} We adopt a standard tagging architecture as our baseline, consisting of an XLM encoder followed by a FFN to predict output labels. For tasks naturally formulated as sequence labeling, this architecture serves as the primary benchmark. However, for parsing tasks, we additionally include established paradigms; specifically, we compare against the span-based model proposed by \citet{kitaev-etal-2019-multilingual} and the biaffine parser \cite{dozat-manning-2017-deep}.

\section{Analysis of results}
Tables~\ref{tab:pos-results} and \ref{tab:ner-results} show PoS and NER accuracies, with relatively small but still unbalanced
label sets
 and weak hierarchies. Tables~\ref{tab:con-results}, \ref{tab:dep-results} and \ref{tab:graph-results} cover constituent,  dependency and graph parsing, respectively.

Table \ref{tab:pos-results} breaks down the PoS tagging results using the universal, language-specific (if available), and rich PoS tags of the UD and SPMRL treebanks. The first two columns summarize the performance of the (x)LSTM-based (non-pretrained) models: although the xLSTM outperforms the LSTM in only 6 out of 22 experiments, the BixLSTM surpasses the BiLSTM in 19 out of 22 PoS tagging tasks. Overall, the BixLSTM is the best non-pretrained encoder in 15 out of 22 experiments, showing
an improvement over its predecessor for simple tagging tasks, although it still falls short of the baseline. Among the pretrained models (DiT, GaT, SSD), the adversarial tagger achieves the best scores in 20 out of 22 experiments (DiT is selected only for the universal German tag, and xLSTM for the rich Hungarian tag), and is the only model to offer competitive performance against the baseline architecture as the top tagger in 9 out of 22 experiments, consistently trailing on others by less than 1 accuracy point.

\begin{table}[h!]\centering\footnotesize 
    \setlength{\tabcolsep}{4pt}
    \renewcommand{\arraystretch}{1}
    \begin{tabular}{|c|c|cc|ccc|c|}
        \cline{3-8}
         \multicolumn{2}{c|}{}&  \multicolumn{2}{c|}{\textbf{LSTM}} & \textbf{DiT} & \textbf{GaT} & \textbf{SSD} & \textbf{Base}\\
         \hline
         \multicolumn{2}{|c|}{PTB} & 95.97 & 96.92\SP{\xlstm} & 95.83 & \textbf{97.91} & 97.33 & \underline{97.97} \\
         \multicolumn{2}{|c|}{CTB} & 92.71 & 93.55\SP{\xlstm} & 97.10 & \underline{\textbf{97.96}} & 93.38 & 97.79 \\
         \hline 
        \hline 
         \parbox[t]{2mm}{\multirow{3}{*}{\textit{de}}}
            & U & 90.31 & 91.52 & \textbf{96.62} & 96.21 & 93.16 & \underline{96.79} \\
            & X & 89.99 & 92.26\SP{\xlstm} & 94.62 & \textbf{97.42} & 92.67 & \underline{97.53} \\
            & R &  68.84\SP{\xlstm} & 71.48\SP{\xlstm} & 65.22 & \underline{\textbf{79.54}} & 65.30 & 78.88 \\
         \hline 
         \parbox[t]{2mm}{\multirow{2}{*}{\textit{eu}}}
            & U &  87.65 & 85.56\SP{\xlstm} & 93.54 & \underline{\textbf{95.54}} & 88.12 & 95.52 \\
            & R &  70.26\SP{\xlstm} & 71.21\SP{\xlstm} & 54.07 & \textbf{72.99} & 59.47 & \underline{74.14} \\
        \hline 
         \parbox[t]{2mm}{\multirow{2}{*}{\textit{fr}}}
            & U &  94.81 & 95.62\SP{\xlstm} & 98.09 & \textbf{98.42} & 96.21 & \underline{98.45} \\
            & R & 80.12\SP{\xlstm} & 85.55\SP{\xlstm} & 78.08 & \underline{\textbf{88.60}} & 80.05 & 88.55 \\
        \hline 
         \parbox[t]{2mm}{\multirow{2}{*}{\textit{he}}}
            & U & 92.47 & 92.05\SP{\xlstm} & 95.24 & \underline{\textbf{97.88}} & 85.34 & 97.52 \\
            & R & 84.43 & 84.54 & 90.98 & \underline{\textbf{95.36}} & 69.11 & \underline{95.36} \\
        \hline 
         \parbox[t]{2mm}{\multirow{2}{*}{\textit{hu}}}
            & U & 77.75 & 75.71 & 95.08 & \textbf{95.81} & 76.42 & \underline{96.28} \\
            & R & \textbf{75.63}\SP{\xlstm} & 71.05\SP{\xlstm} & 61.17 & 74.28 & 60.05 & \underline{75.71} \\
        \hline
         \parbox[t]{2mm}{\multirow{3}{*}{\textit{ko}}}
            & U & 79.89 & 80.54\SP{\xlstm} & 91.14 & \underline{\textbf{94.67}} & 65.05 & 94.49 \\
            & X & 66.57 & 67.10\SP{\xlstm} & 81.40 & \textbf{85.33} & 41.17 & \underline{85.67} \\
            & R &65.24\SP{\xlstm} & 65.52\SP{\xlstm} & 46.65 & \underline{\textbf{66.56}} & 33.29 & 66.42 \\
        \hline 
         \parbox[t]{2mm}{\multirow{3}{*}{\textit{pl}}}
            & U &90.35 & 91.09\SP{\xlstm} & 97.53 & \textbf{98.82} & 90.84 & \underline{98.86} \\
            & X &  76.23 & 79.47\SP{\xlstm} & 92.58 & \underline{\textbf{96.13}} & 71.47 & 95.54 \\
            & R & 64.70\SP{\xlstm} & 64.34\SP{\xlstm} & 48.55 & \textbf{66.10} & 53.52 & \underline{66.92} \\
        \hline 
         \parbox[t]{2mm}{\multirow{3}{*}{\textit{sv}}}
            & U & 88.74 & 90.80\SP{\xlstm} & 97.67 & \textbf{98.18} & 87.59 & \underline{98.22} \\
            & X &83.39 & 82.32\SP{\xlstm} & 92.14 & \underline{\textbf{95.89}} & 78.58 & 95.87 \\
            & R & 80.51 & 80.08\SP{\xlstm} & 90.24 & \textbf{94.38} & 77.19 & \underline{94.61} \\
        \hline 
        \parbox[t]{2mm}{\multirow{3}{*}{$\mu$}} 
            & U & 87.70 & 87.86 & 95.61 & \textbf{96.94} & 87.34 & \underline{97.02} \\
            & X & 79.05 & 80.29 & 90.19 & \underline{\textbf{93.69}} & 70.97 & \underline{93.69} \\
            & R & 73.71 & 74.22 & 66.87 & \textbf{79.73} & 62.25 &  \underline{80.07} \\
        \hline 
    \end{tabular}
    \caption{\label{tab:pos-results}PoS tagging accuracy. Each subrow shows the prediction of a different tag: universal (U) or language-specific (X) for UD; and rich (R) for SPMRL.  \textbf{DiT}, \textbf{GaT} and \textbf{SSD} stand for the diffusion, adversarial and SSD-based models. 
    The \textbf{LSTM} column indicates the best undirectional (first subcolumn) and bidirectional (second subcolumn) LSTM-based model. The \xlstm\ symbol indicates whether the best result comes from the xLSTM. Language acronyms from ISO-639. \textbf{Bold} for the best non-baseline, \underline{underline} for the best SL model. Average across SPMRL and UD treebanks in the last block ($\mu$).}
\end{table}

\begin{table}[tpb!]\centering\footnotesize 
    \setlength{\tabcolsep}{3pt}
    \renewcommand{\arraystretch}{1}
    \begin{tabular}{|c|c|cc|ccc|c|}
        \cline{3-8}
        \multicolumn{2}{c|}{}&  \multicolumn{2}{c|}{\textbf{LSTM}} & \textbf{DiT} & \textbf{GaT} & \textbf{SSD} & \textbf{Base}\\
        \hline
        \multicolumn{2}{|r|}{\scriptsize CoNLL\SPSB{35}{\textit{en}}} & 51.62\SP{\xlstm} & 56.03\SP{\xlstm} & 65.57 & \underline{\textbf{72.63}} & 64.64 & 71.79 \\
        \multicolumn{2}{|r|}{\scriptsize BioNLP\SPSB{39}{\textit{en}}} & 40.90 & 43.88\SP{\xlstm} & 51.62 & \textbf{63.05} & 53.97 & \underline{63.48} \\
        \multicolumn{2}{|r|}{\scriptsize DrugsNER\SPSB{25}{\textit{en}}} & 98.40 & 98.35\SP{\xlstm} & 94.88 & 98.39 & \textbf{98.44} & \underline{98.46} \\
        \multicolumn{2}{|r|}{\scriptsize EIEC\SPSB{26}{\textit{eu}}} & 17.68 & 14.72\SP{\xlstm} & 34.42 & \underline{\textbf{43.61}} & 28.62 & 42.80\\ 
        \multicolumn{2}{|r|}{\scriptsize GermEval\SPSB{16}{\textit{de}}} & 30.64& 28.54 & 44.66 & \textbf{50.03} & 41.95 & \underline{50.16} \\
        \multicolumn{2}{|r|}{\scriptsize HiNER\SPSB{27}{\textit{hi}}} &         67.85 & 71.88\SP{\xlstm} & 69.80 & \textbf{75.25} & 53.43 & \underline{75.30} \\
        \multicolumn{2}{|r|}{\scriptsize JNLPBA\SPSB{40}{\textit{en}}} &  44.42 & 54.46 & 48.46 & \underline{\textbf{62.05}} & 49.30 & 61.99 \\
        \multicolumn{2}{|r|}{\scriptsize NYTK\SPSB{28}{\textit{hu}}} & 39.46 & 37.27\SP{\xlstm} & 71.49 & \textbf{76.32} & 62.80 & \underline{76.36}\\
        \multicolumn{2}{|r|}{\scriptsize Webbnyh.\SPSB{14}{\textit{sv}}} & 20.58 & 19.36 & 29.35 & \textbf{31.39} & 28.02 & \underline{32.58} \\
        \multicolumn{2}{|r|}{\scriptsize Weibo\SPSB{14}{\textit{zh}}} & 21.40\SP{\xlstm} & 26.88\SP{\xlstm} & 19.59 & \underline{\textbf{42.29}}  & 24.65 & 38.58 \\
        \multicolumn{2}{|r|}{\scriptsize WikiNER\SPSB{37}{\textit{fr}}} & 81.71\SP{\xlstm} & 85.59\SP{\xlstm} & 89.72 & \underline{\textbf{92.69}} & 87.32 &  92.63 \\ 
        \multicolumn{2}{|r|}{\scriptsize WikiNER\SPSB{47}{\textit{pl}}} & 82.15 & 84.95 & 86.63 & \underline{\textbf{93.63}} & 87.10 & 93.47 \\
        \hline 
        \multicolumn{2}{|c|}{$\mu$} & 49.73 & 51.82 & 58.84 & \underline{\textbf{66.78}} & 56.68 & 66.46 \\
        \hline 
        \end{tabular}
        \caption{\label{tab:ner-results}Mention-level F1 score. To compare the label set balance, the Shannon entropy is annotated in superscripts. Same notation as in Table \ref{tab:pos-results}.}
\end{table}

Table \ref{tab:ner-results} shows the mention-level F1 score on the NER datasets, where the adversarial tagger 
offers competitive performance against the baseline model, reaching the highest score in 6 out of 12 datasets, while 
\textsc{Mamba-2} achieves the best result on DrugsNER. We hypothesize that due to its reliance on MSE loss, the diffusion tagger underperforms on highly unbalanced datasets like Weibo, failing to surpass even basic non-pretrained models.

Tables~\ref{tab:con-results}, \ref{tab:dep-results} and \ref{tab:graph-results} focus on constituent, dependency and graph parsing. In contrast to structurally simpler tasks (where \textsc{Mamba-2} offered fair performance) our results suggest
that structured state-space models struggle to capture token dependencies within the input sentence, performing on par with the left-to-right (x)LSTM encoder and lagging behind all pretrained models. While the diffusion model achieves relatively good results, it still falls short of the baseline, especially for graph encodings (74.63 vs. 86.33 average LF) under low-resource settings (12.44 vs. 63.99 LF on Tamil). The adversarial tagger slightly improves over the baseline in dependency and constituency parsing: 87.70 vs. 87.49 LAS and 91.04 vs. 90.77 LF, respectively; but matches the baseline in graph parsing (86.27 vs. 86.33 LF).

\begin{table}[h!]\centering\fontsize{9}{9}\selectfont
    \setlength{\tabcolsep}{2pt}
    \renewcommand{\arraystretch}{1.1}
    \begin{tabular}{|c|cc|ccc|c||c|}
        \cline{2-8}
         \multicolumn{1}{c|}{} & \multicolumn{2}{c|}{\textbf{LSTM}} & \textbf{DiT} & \textbf{GaT} & \textbf{SSD} & \textbf{Base} & \textbf{Span} \\
        \hline 
        {\scriptsize PTB} & 59.36\SPSB{\xlstm}{R} & 70.68\SPSB{\xlstm}{R} & 92.76\SB{R} &\underline{\textbf{94.96}}\SB{R} & 63.16\SB{R} & 94.88\SB{R} & \baseline{95.59} \\
        {\scriptsize CTB} & 56.33\SPSB{\xlstm}{T} & 79.64\SPSB{\xlstm}{R} & 87.87\SB{T} & \underline{\textbf{93.52}}\SB{T} & 57.70\SB{T} & 90.84\SB{R} & \baseline{91.75} \\
        \hline 
        \textit{de} & 40.36\SPSB{\xlstm}{T} & 77.29\SPSB{\xlstm}{T} & 88.05\SB{R} & \textbf{91.26}\SB{T} & 40.49\SB{R} & \underline{92.12}\SB{T}& \baseline{90.20} \\
        \textit{eu} & 55.81\SPSB{\xlstm}{T} & 74.62\SPSB{\xlstm}{T} & 80.20\SB{R} & \textbf{89.24}\SB{R} & 48.55\SB{R} & \underline{89.29}\SB{T} & \baseline{91.63}\\
        \textit{fr} & 46.66\SPSB{\xlstm}{T} & 72.63\SPSB{\xlstm}{T} & 84.91\SB{R} & \underline{\textbf{86.52}}\SB{R} & 47.44\SB{R} & 84.79\SB{T}& \baseline{87.42}\\
        \textit{he} & 79.53\SB{T} & 85.25\SPSB{\xlstm}{T} &  91.17\SB{R}  & \underline{\textbf{92.47}}\SB{R} & 61.61\SB{R} & 92.43\SB{R} & \baseline{92.99} \\
        \textit{hu} &  60.32\SB{R} & 71.83\SPSB{\xlstm}{T} & 89.40\SB{R} & \textbf{92.42}\SB{R} & 55.78\SB{R} & \underline{92.47}\SB{R}  & \baseline{94.90}\\
        \textit{ko} & 56.32\SPSB{\xlstm}{T} & 71.00\SPSB{\xlstm}{T} &  85.18\SB{R}  & \underline{\textbf{87.60}}\SB{R} & 41.18\SB{R} & 86.93\SB{T} & \baseline{88.80} \\
        \textit{pl} & 73.19\SPSB{\xlstm}{T} & 89.47\SPSB{\xlstm}{T} & 94.07\SB{T} & \textbf{94.39}\SB{R} & 62.77\SB{R} & \underline{95.85}\SB{T}  & \baseline{96.36}\\
        \textit{sv} &  48.81\SB{R} & 63.03\SPSB{\xlstm}{R} &  81.26\SB{R}  & \textbf{88.02}\SB{R} & 44.68\SB{R} & \underline{88.12}\SB{R} & \baseline{88.87}\\
        \hline 
        $\mu$ & 57.67 & 75.54 & 87.59 & \underline{\textbf{91.04}} & 52.24 & 90.77 & \baseline{91.85}\\
        \hline 
    \end{tabular}
    \caption{\label{tab:con-results}LF score for constituency parsers. Same notation as in Table \ref{tab:pos-results}. Only the best encoding is displayed in subscripts: relative (\textbf{R}) and tetra-tagging (\textbf{T}). The SL (\textbf{Base}) and span-based (\textbf{Span}) baselines are included in the last columns.}
\end{table}

\begin{table}[h!]\centering\fontsize{8}{9}\selectfont
    \setlength{\tabcolsep}{2pt}
    \renewcommand{\arraystretch}{1}
    \begin{tabular}{|c|cc|ccc|c||c|}
        \cline{2-8}
         \multicolumn{1}{c|}{} & \multicolumn{2}{c|}{\textbf{LSTM}} & \textbf{DiT} & \textbf{GaT} & \textbf{SSD} & \textbf{Base} & \textbf{Biaf}\\
        \hline 
        {\scriptsize PTB} & 62.65\SPSB{\xlstm}{B7} & 90.41\SPSB{\xlstm}{B} & 92.39\SB{B7} & \textbf{94.19}\SB{B7} & 66.25\SB{B} & \underline{94.34}\SB{B4} & \baseline{95.36} \\
        {\scriptsize CTB} & 50.41\SPSB{\xlstm}{B} & 82.18\SPSB{\xlstm}{B} & 87.93\SB{B4} & \underline{\textbf{89.19}}\SB{B4} & 44.78\SB{B7} & 88.63\SB{B4} & \baseline{89.67}\\
        \hline 
        \textit{de} & 60.81\SPSB{\xlstm}{B} & 76.07\SPSB{\xlstm}{B7} & 80.75\SB{B4} & \underline{\textbf{83.14}}\SB{B7} & 52.34\SB{B} & 82.77\SB{B7} & \baseline{85.59} \\
        \textit{eu} & 45.48\SPSB{\xlstm}{B} & 68.44\SPSB{\xlstm}{B7} & 79.43\SB{B4} & \textbf{82.54}\SB{B7} & 31.34\SB{B4} & \underline{82.58}\SB{B7} & \baseline{87.35}\\
        \textit{fr} & 69.62\SPSB{\xlstm}{B} & 84.90\SPSB{\xlstm}{H} & 90.78\SB{H} & \underline{\textbf{91.99}}\SB{B} & 67.39\SB{B7} & 91.59\SB{B7} & \baseline{94.14}\\
        \textit{he} & 66.75\SPSB{\xlstm}{B} & 80.31\SPSB{\xlstm}{B7} & 86.43\SB{B7} & \underline{\textbf{89.55}}\SB{B4} & 49.97\SB{B7} & 89.26\SB{H}& \baseline{91.34}\\
        \textit{hu}  & 43.54\SPSB{\xlstm}{B} & 55.05\SPSB{\xlstm}{B4} & 72.16\SB{B4} & \textbf{79.51}\SB{B4} & 23.19\SB{B4} & \underline{80.12}\SB{B7} & \baseline{86.46}\\
        \textit{ko}  & 55.04\SPSB{\xlstm}{B} & 72.85\SPSB{\xlstm}{B4} & 81.52\SB{H} & \underline{\textbf{84.28}}\SB{B} & 22.52\SB{B7} & 83.63\SB{B} & \baseline{83.07} \\
        \textit{pl} & 64.15\SPSB{\xlstm}{B} & 81.04\SPSB{\xlstm}{B4} & 89.30\SB{B7} & \underline{\textbf{91.01}}\SB{B} & 52.98\SB{B} & 90.76\SB{B7} & \baseline{93.85}\\
        \textit{sv}  & 60.32\SPSB{\xlstm}{B4} & 78.06\SPSB{\xlstm}{B4} & 89.54\SB{B4} & \underline{\textbf{91.60}}\SB{B7} & 40.06\SB{B7} & 91.20\SB{B7} & \baseline{91.76}\\
        \hline 
        $\mu$ & 57.88 & 76.93 & 85.02 & \textbf{87.70} & 45.08 & 87.49 & \baseline{89.85}\\
        \hline 
    \end{tabular}
    \caption{\label{tab:dep-results}LAS score for dependency parsers. Same notation as in Table \ref{tab:con-results} with acronyms: 2-planar bracketing (\textbf{B}), 4-bit (\textbf{B4}), 7-bit (\textbf{B7}), hexa-tagging (\textbf{H}). The SL (\textbf{Base}) and biaffine (\textbf{Biaf}) baselines are included in the last columns.}
\end{table}

\begin{table}[h!]\centering\fontsize{8}{9}\selectfont
    \setlength{\tabcolsep}{2pt}
    \renewcommand{\arraystretch}{1.1}
    \begin{tabular}{|c|cc|ccc|c||c|}
        \cline{2-8}
        \multicolumn{1}{c|}{}&  \multicolumn{2}{c|}{\textbf{LSTM}} & \textbf{DiT} & \textbf{GaT} & \textbf{SSD} & \textbf{Base} & \textbf{Biaf}\\
        \hline
        \textit{en} & 65.30\SPSB{\xlstm}{R} & 86.70\SPSB{\xlstm}{B} & 85.91\SB{B} & \underline{\textbf{93.57}}\SB{B} & 64.37\SB{R} & 93.48\SB{B} & \baseline{94.15} \\
        \textit{zh} & 46.87\SPSB{\xlstm}{R} & 76.03\SPSB{\xlstm}{$6k$} & 64.09\SB{$6k$} & \textbf{83.00}\SB{$6k$} & 37.64\SB{R} & \underline{83.23}\SB{$6k$}  & \baseline{88.91} \\
        \textit{ar} & 67.86\SPSB{\xlstm}{R} & 73.31\SPSB{\xlstm}{$6k$} & 76.61\SB{$6k$} & \underline{\textbf{81.75}}\SB{$6k$} & 57.09\SB{R} & 81.08\SB{B}  & \baseline{84.74}\\
        \textit{bg} & 67.78\SPSB{\xlstm}{$6k$} & 82.78\SPSB{\xlstm}{$6k$} & 86.83\SB{$6k$} & \textbf{90.47}\SB{$4k$} & 50.31\SB{$6k$} & \underline{90.64}\SB{$4k$}  & \baseline{93.57} \\
        \textit{cs} & 59.82\SPSB{\xlstm}{$6k$} & 78.95\SPSB{\xlstm}{R} & 82.27\SB{$4k$} & \underline{\textbf{88.45}}\SB{$6k$} & 53.92\SB{$6k$} & 88.23\SB{$6k$}  & \baseline{89.79} \\
        \textit{fr}  & 70.95\SPSB{\xlstm}{$6k$} & 82.21\SPSB{\xlstm}{$6k$} & 83.29\SB{$6k$} & \underline{\textbf{91.73}}\SB{$6k$} & 66.00\SB{$6k$} & 90.97\SB{$6k$}  & \baseline{95.10} \\
        \textit{it} & 72.83\SPSB{\xlstm}{$6k$} & 86.69\SPSB{\xlstm}{$6k$} & 87.85\SB{$6k$} & \textbf{91.46}\SB{$4k$} & 69.46\SB{$6k$} & \underline{91.87}\SB{$6k$} & \baseline{94.00} \\
        \textit{nl}  & 59.30\SPSB{\xlstm}{$6k$} & 77.19\SPSB{\xlstm}{R} & 81.70\SB{$6k$} & \textbf{87.75}\SB{$6k$} & 48.74\SB{$6k$} & \underline{87.88}\SB{$6k$}  & \baseline{94.38} \\
        \textit{pl} & 64.06\SPSB{\xlstm}{$6k$} & 81.38\SPSB{\xlstm}{R} & 85.27\SB{$6k$} & \textbf{91.76}\SB{$6k$} & 54.17\SB{$6k$} & \underline{91.97}\SB{$6k$}  & \baseline{92.08} \\
        \textit{ta} & 40.49\SPSB{\xlstm}{$6k$} & 52.21\SPSB{\xlstm}{$4k$} & 12.44\SB{B} & \textbf{62.79}\SB{$6k$} & 15.81\SB{R} & \underline{63.99}\SB{$6k$}  & \baseline{92.06} \\
        \hline 
        $\mu$ & 61.53 & 77.74 & 74.63 & \textbf{86.27} & 51.75 & \underline{86.33}  & \baseline{91.87} \\
        \hline 
    \end{tabular}
    \caption{\label{tab:graph-results}LF score for graph parsers. Same notation as in Tables \ref{tab:con-results} and \ref{tab:dep-results}, with acronyms: relative (\textbf{R}), bracketing with $k=3$ (\textbf{B}), $4k$-bit with $k=4$ ($\boldsymbol{4k}$) and $6k$-bit with $k=4$ ($\boldsymbol{6k}$) encodings. The SL (\textbf{Base}) and biaffine (\textbf{Biaf}) baselines are included in the last columns.}
\end{table}

To assess the gains of the adversarial tagger, we measured the ratio of well-formed label sequences
\footnote{We consider a label sequence to be well-formed if it corresponds to the encoding of a tree, that is, if a valid tree is obtained directly from the decoding process directly, without the need of postprocessing to deal with issues like undefined nonterminals or out-of-range indexes.}
on the PTB constituency treebank (chosen due to space reasons and its widespread use). The adversarial tagger produced valid label sequences for 82.13\% of the test set, compared to 80.76\% for the baseline. 
This likely stems from the adversarial loss, which pushes the tagger to produce sequences that fool the discriminator, improving performance. The effect weakens in graph parsing, where label sequences are less constrained, as graphs are more expressive and need not be connected or acyclic.

\begin{figure}[h!]
    \includegraphics[width=\linewidth]{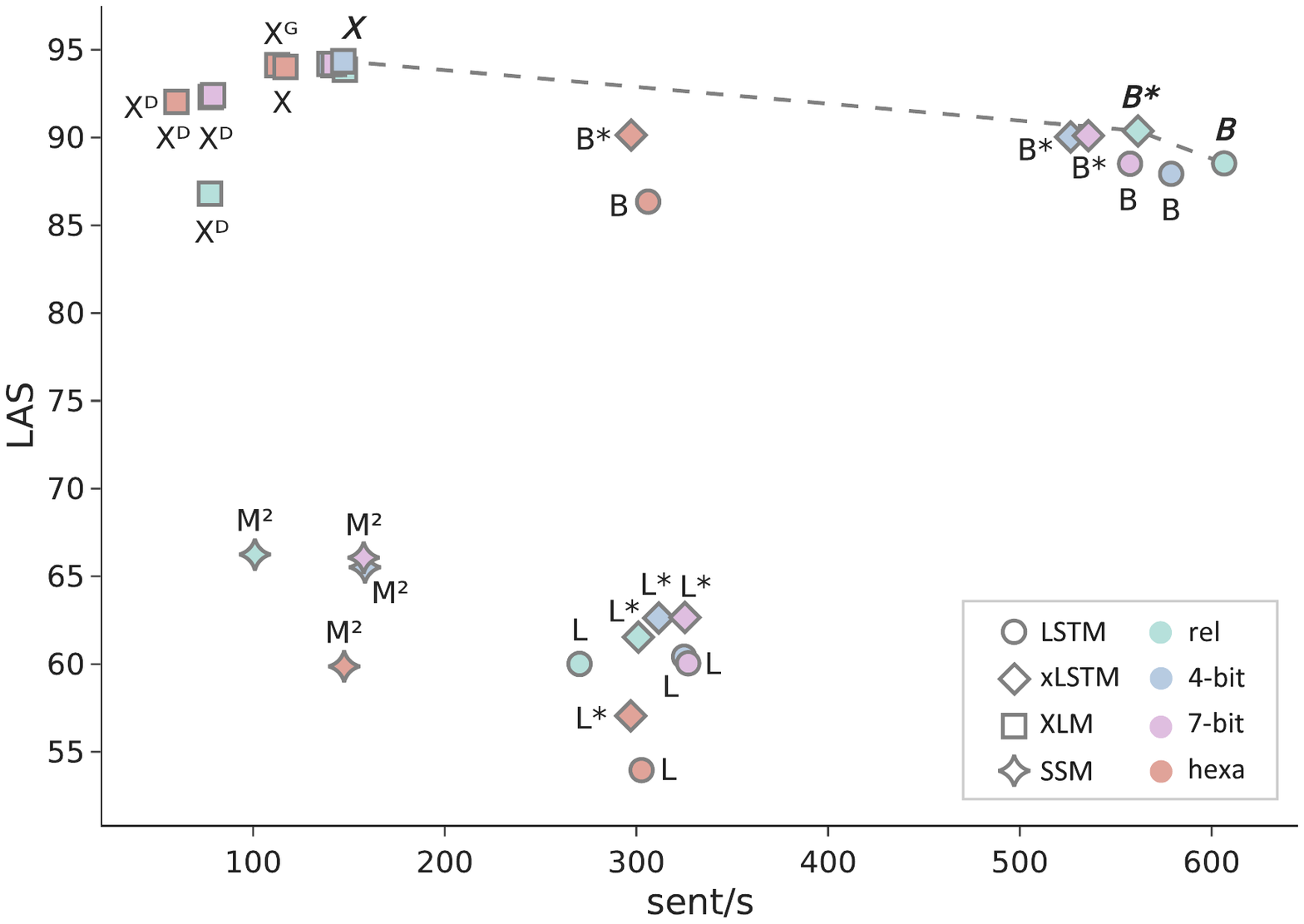}
    \caption{\label{fig:pareto}
    Pareto front of LAS vs. speed (sent/s) on PTB dependency parsing.
    Colors are reserved for encodings, symbols and text annotations for architectures: LSTM (\textbf{L}), BiLSTM (\textbf{B}), xLSTM (\textbf{L*}), BixLSTM (\textbf{B*}), \textsc{Mamba-2} (\textbf{M\textsuperscript{2}}), XLM (\textbf{X}), DiT (\textbf{X\textsuperscript{D}}) and GaT (\textbf{X\textsuperscript{G}}). All executions were measured on the same NVIDIA RTX 3090 24GB.
    }
\end{figure}

\paragraph{Speed analysis} Figure \ref{fig:pareto} compares the performance vs speed of our dependency parsers. Unidirectional encoders (LSTM, xLSTM, \textsc{Mamba}) although performing well on tagging, suffer considerably in parsing. The baseline and adversarial tagger pair in speed since at inference time both rely on the same number of parameters. The BiLSTM and BixLSTM, although falling behind the pretrained models by $\sim$4 points, offer a fast yet accurate approach.\footnote{In our LSTM-based encoder, each layer is a LSTM  unit with hidden size  $d=400$. For their bidirectional counterpart  each layer has two units---each processing the input in a different directions---with $d=200$, maintaining the overall layer size to $400$. Under this configuration the complexity of the unidirectional  encoder scales with $d=400$  while the bidirectional  with $d=200$.}

\begin{figure}[h!]\centering
    \includegraphics[width=\linewidth]{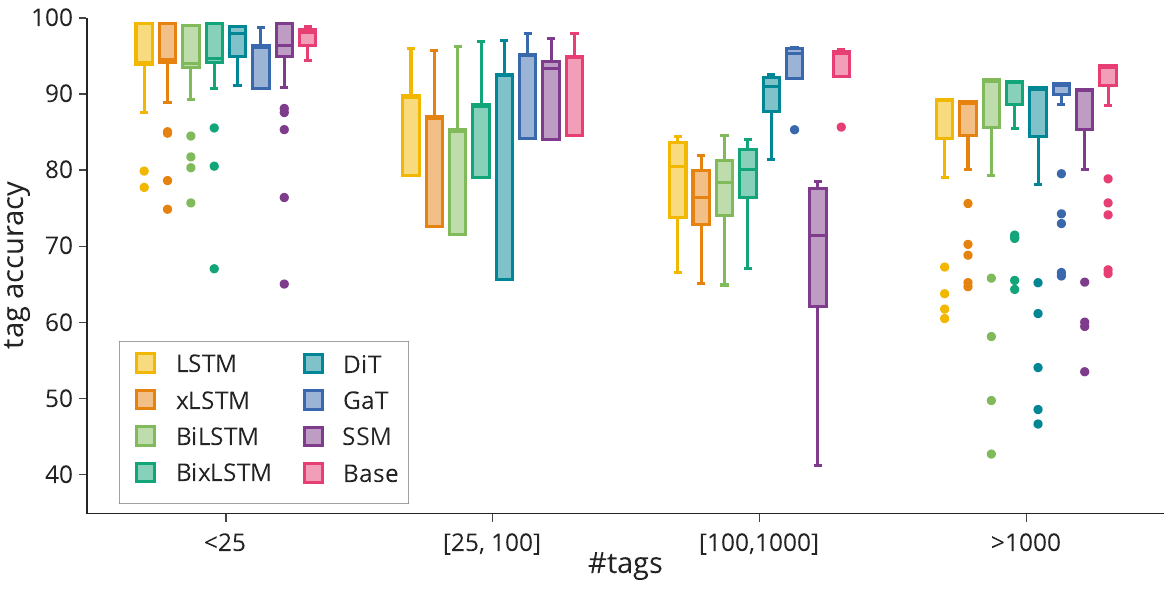}
    \caption{\label{fig:tag-boxplot}
    PoS and NER accuracy ($y$) across output spaces ($x$, uneven intervals).
    }
\end{figure}

\paragraph{Impact of label space} Figure~\ref{fig:tag-boxplot} shows the performance consistency of each architecture as the label space increases in PoS tagging and NER tasks. In general, all models experience a degradation in performance as the output space grows, particularly non-Transformer models (LSTM- and SSD-based), which exhibit greater performance variability even in smaller label spaces. 
Among non-baselines, the adversarial tagger is most stable, with few outliers even under large label spaces, and performance matching the baseline.

\begin{figure}[h!]
    \hspace{-1em}\includegraphics[width=1.05\linewidth]{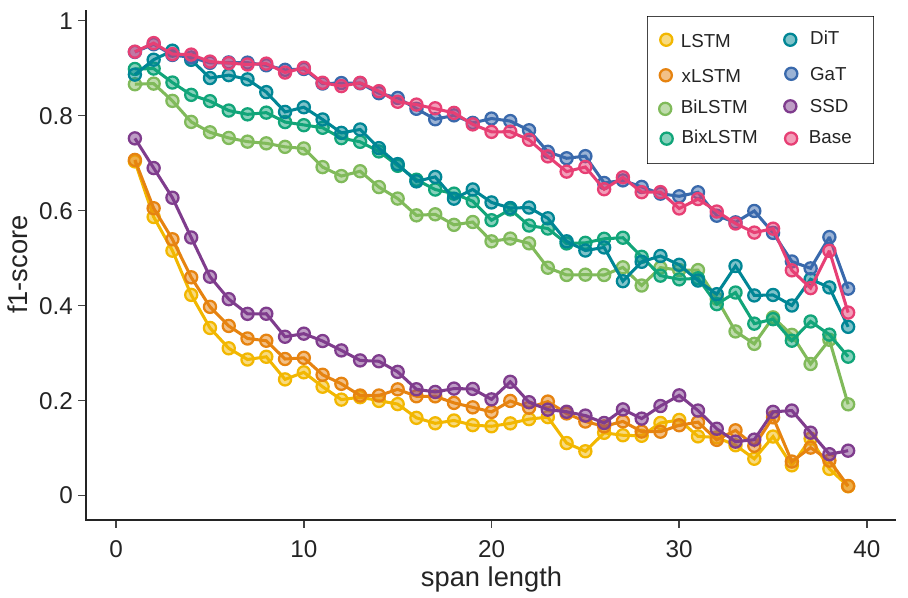}
    
    \hspace{-1em}\includegraphics[width=1.05\linewidth]{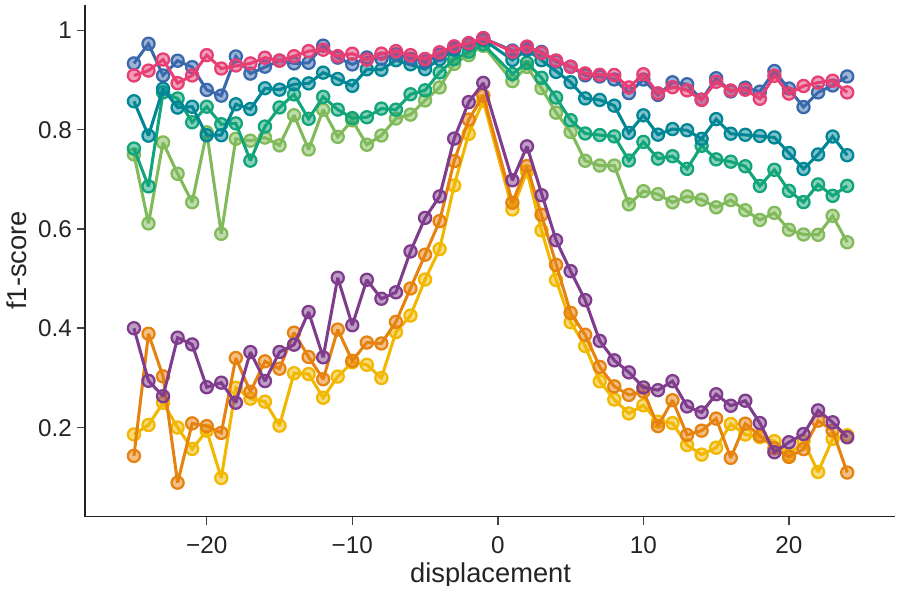}
    \caption{\label{fig:ptb-spans-displ}
    Parsing performance on the PTB test set. The top panel shows F1-scores relative to span length (constituent parsing), while the bottom panel shows performance relative to dependency displacement (dependency parsing).}
    
\end{figure}

\paragraph{Parsing performance} Figure \ref{fig:ptb-spans-displ} provides a fine-grained analysis of parsing performance on the PTB dataset, specifically evaluating F1-scores relative to constituent span length and dependency displacement.\footnote{Following \citet{anderson-gomez-rodriguez-2020-inherent}, we define displacement as the signed distance $d-h$ between the dependent ($d$) and head ($h$) positions.} Broadly, 
unidirectional models (LSTMs and state-space models) exhibit a substantial
performance decay when modeling long-range dependencies compared to their bidirectional counterparts. This trend aligns with recent findings on the challenges of incrementality \cite{ezquerro-etal-2023-challenges, EzqGomVilEACL2024}, highlighting a persistent limitation that remains unresolved even in large pretrained architectures like \textsc{Mamba}. Additionally, while the diffusion tagger shows promise on simpler tagging problems (Table \ref{tab:pos-results}), it barely outperforms 
Bi(x)LSTM encoders on parsing tasks, despite
relying on a powerful Transformer-based encoder. This marginal gain suggests that diffusion tagging still struggles to effectively internalize hierarchical information as a sequence of tags.

\section{Conclusion}
In this work we study alternative architectures for sequence labeling, examining models such as diffusion and adversarial approaches, which fully redefine how the label space is modeled, and non-Transformer contextualizers like xLSTMs and SSMs, which modify the underlying backbone for sequence modeling. Our unified evaluation reveals that both lines of work can rival or even surpass standard Transformer-based setups on simple tasks such as PoS tagging and NER. However, when considering more complex tasks that demand long-range dependencies, only the adversarial tagger maintains competitive performance against traditional methods. Its ability to exploit generative modeling proves especially effective in capturing well-formed structures beyond the capabilities of other non-standard architectures.

\section*{Acknowledgments}

We acknowledge grants GAP (PID2022-139308OA-I00) funded by MICIU/AEI/10.13039/501100011033/ and ERDF, EU; LATCHING (PID2023-147129OB-C21) funded by MICIU/AEI/10.13039/501100011033 and ERDF, EU; and TSI-100925-2023-1 funded by Ministry for Digital Transformation and Civil Service and ``NextGenerationEU'' PRTR; as well as funding by Xunta de Galicia (ED431C 2024/02), and CITIC, as a center accredited for excellence within the Galician University System and a member of the CIGUS Network, receives subsidies from the Department of Education, Science, Universities, and Vocational Training of the Xunta de Galicia. Additionally, it is co-financed by the EU through the FEDER Galicia 2021-27 operational program (Ref. ED431G 2023/01). This research project was made possible through the access granted by the Galician Supercomputing Center (CESGA) to its supercomputing infrastructure. The supercomputer FinisTerrae III and its permanent data storage system have been funded by the NextGeneration EU 2021 Recovery, Transformation and Resilience Plan, ICT2021-006904, and also from the Pluriregional Operational Programme of Spain 2014-2020 of the European Regional Development Fund (ERDF), ICTS-2019-02-CESGA-3, and from the State Programme for the Promotion of Scientific and Technical Research of Excellence of the State Plan for Scientific and Technical Research and Innovation 2013-2016 State subprogramme for scientific and technical infrastructures and equipment of ERDF, CESG15-DE-3114.

\section*{Limitations} 
\paragraph{Physical resources} Our computational resources consists of 8 24GB RTX 3090 and 3 40GB A100 GPUs. We also have limited access to a large computing infrastructure with more than 300 nodes, where each node contains 8 40GB A100. 

\paragraph{Lack of generative models as encoders}  
For models that include Transformers as components, we rely on masked language models (MLMs) rather than generative models. The first reason is that incremental sequence tagging with autoregressive models lags behind bidirectional encoders, as recently shown by \citet{ezquerro-etal-2023-challenges,EzqGomVilEACL2024}. Given this, our choice of masked language models aligns with current best practices for sequence labeling. The second reason is computational constraints. While our setup allowed extensive experimentation, training large-scale generative models would require significantly higher resources. As such, our focus remains on models that are both effective and computationally accessible.

\bibliography{custom}

\appendix
\section{Visualizations}\label{ap:figures}
\begin{figure*}[htpb!]\centering
    \begin{subfigure}[t]{0.51\textwidth}
        \caption{\label{fig:diffusion-training}Forward process.}
        \includegraphics[width=\textwidth]{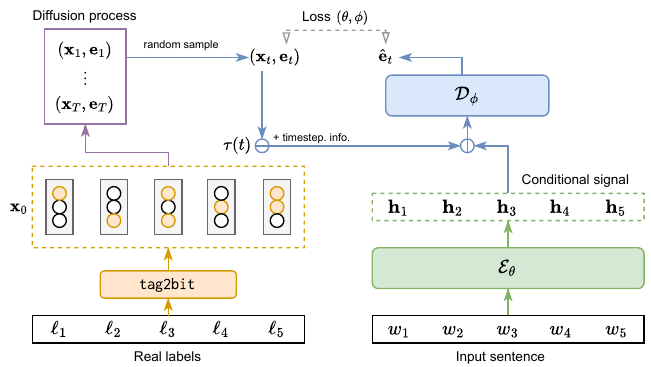}
    \end{subfigure}
    \hfill
    \begin{subfigure}[t]{0.48\textwidth}
        \caption{\label{fig:diffusion-inference}Denoising process.}
        \includegraphics[width=\textwidth]{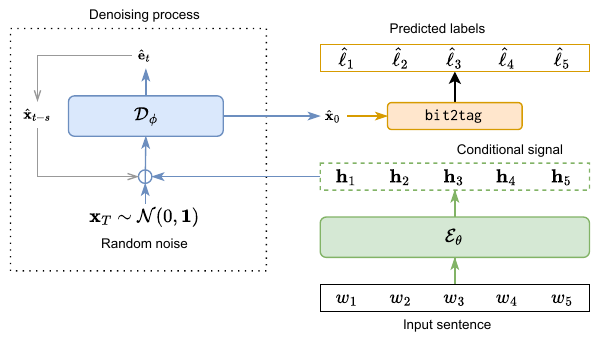}
    \end{subfigure}
    \caption[Diffusion tagger]{\label{fig:diffusion}
    Diffusion tagger in forward and denoising steps. The symbol \rotatebox[origin=c]{90}{$\ominus$} is the concatenation operator and an open arrow (\DiagonalArrow) loss propagation. In Figure~\ref{fig:diffusion-training}, $\mathcal{E}_\theta$ embeds the sentence as the conditional signal. The real labels are transformed into bits and fed to the diffusion process, where the latent $\mathbf{x}_t$ is computed from the sampled noise $\mathbf{e}_t$, and concatenated with time embeddings $\tau(t)$ and the conditional signal. Then, $\mathcal{D}_\phi$ learns to extract the noise that was added to $\mathbf{x}_t$. All parameters are optimized with the MSE loss between the real and predicted noise. Figure \ref{fig:diffusion-inference} shows the denoising process. The conditional signal is computed once with $\mathcal{E}_\theta$ and an initial signal $\mathbf{x}_T$ is sampled from Gaussian noise. Iteratively, $\mathcal{D}_\phi$ removes noise from the input and conditional signal and estimates the previous latent $\hat{\mathbf{x}}_{t-s}$ until $\hat{\mathbf{x}}_0$ is reached. Then, $\hat{\mathbf{x}}_0$ is fed to the BT module to recover a sequence of predicted labels.}
\end{figure*}

\begin{figure}[ht!]\centering
    \includegraphics[width=\linewidth]{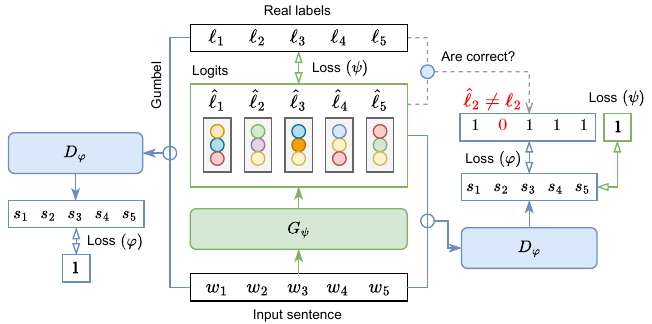}
    \caption{\label{fig:adversarial}Adversarial tagger view (symbols as in Figure~\ref{fig:diffusion}). 
    $G_\psi$ (green) is trained with the tag loss. $D_\varphi$ (blue) learns to distinguish valid tag sequences and guides $G_\psi$.
    }
\end{figure}

Figure \ref{fig:diffusion} illustrates the forward and denoising process of the diffusion tagger. Figure \ref{fig:adversarial} shows the loss flow of the adversarial training.

\section{Treebank statistics}\label{ap:treebanks}

Table \ref{tab:treebanks} summarizes the number of tags required in our datasets. We selected the following UD treebanks to train and evaluate our dependency parsers: German GSD, Basque BDT, French GSD, Hebrew HTB, Hungarian Szeged, Korean KAIST, Polish PDB, and Swedish Talbanken. For graph parsing, we drew from the SDP \cite{oepen-etal-2015-semeval} and IWPT \cite{bouma-etal-2021-raw} datasets, selecting the English DM, Chinese PAS, Arabic PADT, Bulgarian BTB, Czech PSD, French Sequoia, Italian IST, Dutch Alpino, Polish PDB, and Tamil TTB treebanks. These languages were chosen to reflect a broad range of syntactic structures and to overlap with those in the SPMRL corpus, enabling more consistent cross-task comparisons of tagging performance in typologically similar languages.

\begin{table}[htpb!]\centering\scriptsize
    \setlength{\tabcolsep}{1.2pt}
    \renewcommand{\arraystretch}{1.1}
    \begin{minipage}{0.6\linewidth}
    \begin{tabular}{|c|ccc|ccc|cc|}
        \cline{2-9}
        \multicolumn{1}{c|}{}& \multicolumn{6}{c|}{\textbf{Dep.}} & \multicolumn{2}{c|}{\textbf{Cons.}} \\
        \cline{2-9}
        \multicolumn{1}{c|}{}& \textbf{\tiny UPOS} & \textbf{\tiny XPOS} & \textbf{\tiny REL} & \textbf{\tiny B} & \textbf{\tiny B4} & \textbf{\tiny B7} & \textbf{\tiny POS} & \textbf{\tiny R}\\
        \hline 
        PTB & 45 & 45 & 45 & 235 & 16 & 22 & 46 & 39 \\
        CTB &  36 & 36 & 19 & 337 & 16 & 16 & 42 & 26 \\ 
        \hline 
        \textit{de} & 17 & 52 & 45 & 311 & 16 & 69 & 30k & 16 \\ 
        \textit{eu} &  17 & - & 33 & 250 & 16 & 81 & 30k & 16 \\
        \textit{fr} & 16 & 2 & 56 & 235 & 16 & 51 & 39k & 28 \\
        \textit{he} & 15 & - & 37 & 164 & 16 & 37  & 281 & 33 \\
        \textit{hu} & 16 & - & 52 & 196 & 16 & 61 & 50k & 19 \\
        \textit{ko} &  17 & 2k & 32 & 150 & 16 & 40 & 105k & 25 \\
        \textit{pl} & 17 & 856 & 67 & 246 & 16 & 71 & 27k & 21 \\
        \textit{sv} & 17 & 144 & 44 & 176 & 16 & 46 & 357 & 23 \\
        \hline 
    \end{tabular}
    \end{minipage}
    \begin{minipage}{0.38\linewidth}
    \begin{tabular}{|c|cccc|}
        \cline{2-5}
        \multicolumn{1}{c|}{}& \multicolumn{4}{c|}{\textbf{Graph}} \\ 
        \cline{2-5}
        \multicolumn{1}{c|}{}& \textbf{\tiny R} & \textbf{\tiny B} & $\tiny \boldsymbol{4k}$ & $\tiny \boldsymbol{6k}$\\
        \hline 
        \textit{en} & 8139 & 506 & 710 & 650 \\
        \textit{zh} & 28567 & 1315 & 1001 & 859 \\
        \textit{ar} & 2665 & 1076 & 448 & 466 \\
        \textit{bg} & 917 & 522 & 268 & 239 \\
        \textit{cs} & 5200 & 2225 & 1179 & 1139 \\
        \textit{fr} & 968 & 578 & 229 & 208 \\
        \textit{it} & 2004 & 900 & 306 & 285 \\
        \textit{nl} & 1093 & 884 & 371 & 321 \\
        \textit{pl} & 2380 & 1541 & 679 & 598 \\
        \textit{ta} & 194 & 118 & 74 & 45 \\
        \hline 
    \end{tabular}
    \end{minipage}
    \\[1em]
    \begin{tabular}{cccccc}
        \hline 
         {\scriptsize CoNLL} &  {\scriptsize BioNLP} & {\scriptsize DrugsNER} & {\scriptsize EIEC} & {\scriptsize GermEval}   & {\scriptsize HiNER} \\
         \hline 
         45 & 52 & 1611 & 9 & 25 & 23  \\ 
         \hline 
    \end{tabular}\\[1em]
    \begin{tabular}{cccccc}
        \hline 
        {\scriptsize JNLPBA} &  {\scriptsize NYTK} & {\scriptsize Webbnyh.} & {\scriptsize Weibo} & {\scriptsize WikiNER} \\ 
         \hline 
         11 & 9 & 5 & 17 & 4 \\
         \hline 
    \end{tabular}

    \caption{\label{tab:treebanks}Number of learned tags in our experiments for PoS tagging; dependency, constituency and graph parsing; and NER. Languages are specified with the ISO-639 code and the corpus is specified in subscripts. Encodings are abbreviated as in Tables~\ref{tab:con-results}, \ref{tab:dep-results} and \ref{tab:graph-results}: bracketing (\textbf{B}), 4-bit (\textbf{B4}) and 7-bit (\textbf{B7}) encodings for dependency parsing; relative (\textbf{R}) for constituency parsing; and relative (\textbf{R}), bracketing with $k=3$ (\textbf{B}), $4k$-bit with $k=4$ ($\boldsymbol{4k}$) and $6k$-bit with $k=4$ ($\boldsymbol{6k}$). \textbf{UPOS}, \textbf{XPOS} and \textbf{REL} refer to the number of unique tags in the CoNLL format; and \textbf{POS} indicates the number of tags of the PTB, CTB and SPMRL corpus.}
\end{table}

\section{Existing parsing linearizations}\label{ap:examples}
In our empirical study, we relied on existing sequence-labeling approaches to cast tree- or graph-structured prediction tasks in terms of token-level classification. In this section, we provide a detailed description of each encoding, along with illustrative examples to clarify the mapping from the original structured input to its token-level representation. We also highlight the advantages and potential limitations of each approach in preserving critical structural dependencies.

\subsection{Constituency parsing} 
\begin{figure}[h!]\centering
    \begin{subfigure}{\linewidth}
        \caption{\label{fig:con-relative-example}Absolute (\textbf{A}) and relative (\textbf{R}) encodings  \cite{gomez-rodriguez-vilares-2018-constituent}. The row \textbf{Cons.} is the second component of the label and remains the same for both variants.}
        \includegraphics[width=\linewidth]{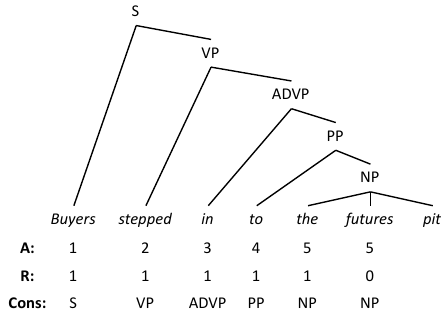}
    \end{subfigure} 
    \begin{subfigure}{\linewidth}
        \caption{\label{fig:con-tetra-example}Tetratagging \cite{kitaev-klein-2020-tetra} for a binarized tree:  \textbf{tags} represent the first component of the label, while \textbf{fences} represent the second component on the fencepost positions.}
        \includegraphics[width=0.95\linewidth]{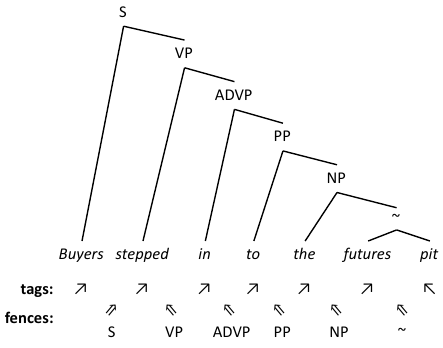}
    \end{subfigure}
    \caption{\label{fig:constituent-tree}Example of a constituent tree encoded with the relative encoding (Figure \ref{fig:con-relative-example}) and tetratagging (Figure \ref{fig:con-tetra-example}).}
\end{figure}

For our study, we relied on the relative \cite{gomez-rodriguez-vilares-2018-constituent}
and tetratagging \cite{kitaev-klein-2020-tetra} linearizations. 
Figure~\ref{fig:constituent-tree} illustrates a constituent tree and its transformation into label sequences under each encoding scheme.

\paragraph{Relative encoding} Given a constituent tree with no unary chains\footnote{As proposed by \citet{gomez-rodriguez-vilares-2018-constituent}, we remove unary chains by collapsing their constituents into a single node (e.g., the chain \texttt{S-NP} is collapsed into a single node with the constituent \texttt{S:NP}).} over the sentence $(w_1\cdots w_n)$, the absolute encoding proposed by \citet{gomez-rodriguez-vilares-2018-constituent} represents an input tree with a sequence of $n-1$ labels, where each label consists of two components $\ell^{A}_i=(p_i, c_i)$, for $i=1,...,n-1$. The first component $p_i\in\mathbb{Z}^+$ indicates the number of constituents shared between $w_i$ and $w_{i+1}$; while the second component $c_i$ is the lowest shared constituent. The relative encoding is directly obtained from the absolute encoding by building each label as $\ell^\text{R}_i=(p_i-p_{i-1}, c_i)$ where $i>1$; otherwise $\ell_i^\text{R}=\ell_i^\text{A}$.

The relative decoding only requires processing the tree from left to right, opening intermediate nodes when upon encountering a positive value in the first component of each label and resolving them when finding negative values or reaching the end of the sequence.

\paragraph{Tetratagging} \citet{kitaev-klein-2020-tetra} encode a binary constituent tree with $n$ labels, where each label $\ell_i^\text{T}=(t_i,f_i,c_i)$ is composed of three components, and the last one is always defined as $\ell_n^\text{T}=(\nwarrow, \emptyset,\emptyset)$. The first component of the label $t_i$ represents with an arrow symbol  whether the terminal node $w_i$ is a left ($\nearrow$) or right descendant ($\nwarrow$) of its parent. The second and third component, $f_i$ and $c_i$, also determine whether the lowest nonterminal node that covers the fencepost between $w_i$ and $w_{i+1}$ is a left ($\Nearrow$) or right ($\Nwarrow$) child and the constituent of the lowest common node.

\subsection{Dependency parsing}
\begin{figure*}[h!]\small\centering
    \begin{subfigure}{\linewidth}\centering
        \caption{\label{fig:projective-tree-example}Projective dependency tree and the bracketing (\textbf{B}) and 4-bit encoding (\textbf{B4}).}
        \begin{dependency}
            \begin{deptext}[column sep=2pt]
                $w_0$ \& \textit{I} \& \textit{had} \& \textit{to} \& \textit{go} \& \textit{to} \& \textit{the} \& \textit{BBC} \& \textit{for} \& \textit{this} \& \textit{report} \\[1em]
                \textbf{B}:\& \texttt{<} \& \texttt{\char`\\>/} \& \texttt{<} \& \texttt{\char`\\>//} \& \texttt{<} \& \texttt{<} \& \texttt{\char`\\\char`\\>} \& \texttt{<} \& \texttt{<} \& \texttt{\char`\\\char`\\>}\\[1em] 
                \textbf{B4}: \& 0100 \& 1111 \& 0100 \& 1111 \& 0100 \& 0000 \& 1010 \& 0100 \& 0000 \& 1110 \\[1em]
                \textbf{REL}: \& nsubj \& root \& mark \& xomp \& case \& det \& obl \& case \& det \& obl\\ 
            \end{deptext}
            \depedge{3}{2}{nsubj}
            \depedge{5}{4}{mark}
            \depedge{3}{5}{xcomp}
            \depedge{8}{6}{case}
            \depedge{8}{7}{det}
            \depedge{5}{8}{obl}
            \depedge{11}{9}{case}
            \depedge{11}{10}{det}
            \depedge[edge height=1.9cm]{5}{11}{obl}
            \depedge{1}{3}{root}
        \end{dependency}
    \end{subfigure}
    \hfill
    \begin{subfigure}{\linewidth}\centering 
        \caption{\label{fig:projective-con-tree-example}Dependency tree of Figure \ref{fig:projective-tree-example} transformed into a binary constituent tree and encoded with tetratagging.}
        \includegraphics[width=0.45\linewidth]{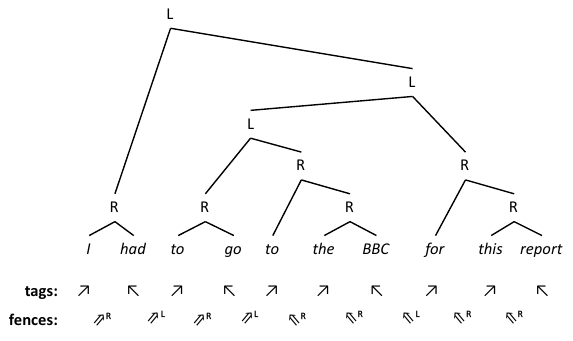}
    \end{subfigure}
    \begin{subfigure}{\linewidth}\centering
        \caption{\label{fig:2-planar-tree-example}2-planar dependency tree encoded with the bracketing (\textbf{B}) and 7-bit encoding (\textbf{B7}).}
        \begin{dependency}
            \begin{deptext}[column sep=2pt]
                $w_0$ \& \textit{Any} \& \textit{particular} \& \textit{shop} \& \textit{that} \& \textit{you} \& \textit{know} \& \textit{of} \& \textit{and} \& \textit{their} \& \textit{number} \\[1em]
                \textbf{B}: \& \texttt{<} \& \texttt{<} \& \texttt{\char`\\\char`\\>//} \& \texttt{\textcolor{BrickRed}{/*}<} \& \texttt{<} \& \texttt{\char`\\\char`\\>} \& \texttt{\textcolor{BrickRed}{>*}} \& \texttt{<} \& \texttt{<} \& \texttt{\char`\\\char`\\>}\\[1em] 
                \textbf{B7}: \&  0010000\& 0000000\& 1011100\&  0010001\& 0000000\&  1001000\&  1110000\&  0010000\&  0000000\&  1001000 \\
            \end{deptext}
            \depedge{4}{2}{det}
    	\depedge{4}{3}{amod}
    	\depedge{7}{5}{obl}
    	\depedge[edge height=0.5cm]{7}{6}{nsubj}
    	\depedge[edge height=1.5cm]{4}{7}{acl:relcl}
    	\depedge[edge height=1.2cm, edge horizontal padding=-1ex, edge style={color=BrickRed}]{5}{8}{case}
    	\depedge[edge height=0.5cm]{11}{10}{cc}
    	\depedge[edge height=1cm]{11}{9}{nmod:poss}
    	\depedge[edge height=2cm]{4}{11}{conj}
    	\depedge{1}{4}{root}
        \end{dependency}
    \end{subfigure}

    \caption{\label{fig:dep-tree}Projective (Figures \ref{fig:projective-tree-example} and \ref{fig:projective-con-tree-example}) 2-planar (Figure \ref{fig:2-planar-tree-example}) dependency tree examples.}
\end{figure*}

Given a dependency tree $G=(W,A)$, where $W=(w_1\cdots w_n)\in\mathcal{V}^n$ is the input sentence, and $A=\{(h\overset{r}{\to} d): d=1\cdots  n; h\in[0,n], h\neq d, r\in\mathcal{R}\}$\footnote{We use the symbol $\mathcal{R}$ to denote the set of arc labels (dependency types).} is the arc set, a tree linearization represents the information of $A$ as a sequence of $n$ labels $(\ell_1 \cdots \ell_n)\in\mathcal{L}^n$. Figure \ref{fig:dep-tree} shows two examples of a dependency tree encoded with the 4-bit and 7-bit encodings \cite{gomez-rodriguez-etal-2023-4} and hexatagging \cite{amini-etal-2023-hexatagging}. 

\paragraph{Projectivity and planarity} A dependency tree $G$ is a connected, acyclic labeled graph where each node has only one incoming arc. We say that a dependency tree is \emph{projective} when no arcs of $A$ crosses each other. For instance, the tree in Figure \ref{fig:projective-tree-example} is projective, while the tree in Figure \ref{fig:2-planar-tree-example} is non-projective because the arc ($4\overset{\text{case}}{\longrightarrow} 7$) crosses $(3\overset{\text{acl:relcl}}{\longrightarrow} 6)$. 

Assuming a non-projective dependency tree, the arcs of $A$ can be distributed into at least $k$ mutually exclusive subsets (also denoted as \emph{planes}) of non-crossing arcs. We say then that the dependency graph is $k$-planar, meaning that the minimum number of planes into which $A$ can be distributed is equal to $k$. The dependency tree displayed in Figure \ref{fig:2-planar-tree-example} is 2-planar, since the crossing arc ($4\overset{\text{case}}{\longrightarrow} 7$) needs to be located in a second plane. 

Projective encodings (4-bit, hexatagging) only recover the full set of arcs when the dependency tree is projective (i.e., 1-planar). Although most of the English treebanks can be covered at $>99\%$ by projective encodings, \citet{amini-etal-2023-hexatagging} used a pseudo-projectivity transformation\footnote{The pseudo-projectivity proposed by \citet{nivre-nilsson-2005-pseudo} is a lossy transformation with three variants (\textit{head}, \textit{path}, and \textit{head+path}) that encodes the information of the crossing arcs in the arc labels---thus increasing the complexity of this label space $\mathcal{R}$.} to train the hexatagger. For our experiments, we also applied pseudo-projectivity to train the parsers with projective tree encodings.

\paragraph{Hexatagging} \citet{amini-etal-2023-hexatagging} encode a dependency tree $G=(W,A)$ with a sequence of labels where each label  $\ell_i=(h_i, f_i, r_i)$ has three components: $h_i\in\{\nearrow, \nwarrow\}$, $f_i\in\{\Nwarrow^\text{R}, \Nwarrow^\text{L}, \Nearrow^\text{R}, \Nearrow^\text{L}\}$ and $r_i\in\mathcal{R}$; constrained to $\ell_1=(\nearrow, f_1, r_1)$ and $\ell_n=(\nwarrow, \Omega, r_n)$. 

The hexatagger first projectivizes a dependency tree \cite{nivre-nilsson-2005-pseudo} and then transforms it into a binary constituent tree (BHT) with constituents in $\mathcal{C}=(\texttt{R},\texttt{L})$ that is encoded with tetratagging \cite{kitaev-klein-2020-tetra}. Figure \ref{fig:projective-con-tree-example} shows the resulting BHT from the projective dependency tree in Figure \ref{fig:projective-tree-example}. The first component of the label $h_i$ corresponds to the first label of tetratagging, and the second component $f_i$ to the concatenated fencepost symbols and their corresponding constituent. The third component is the incoming arc label of $w_i$ (row \textbf{REL} in Figure \ref{fig:projective-tree-example}).

\paragraph{Bracketing encoding} \citet{strzyz-etal-2019-viable} use bracket symbols $B=\{\texttt{/},\texttt{>}, \texttt{<}, \texttt{\char`\\}\}$ to encode the information of the incoming and outgoing arcs of each node. Under the bracketing encoding, each label is defined as $\ell_i=(b_i,r_i)$. The first symbol $b_i$ follows the regular expression \texttt{\char`\\*(>|<)/*} and  the presence of each bracket represents an incoming (\texttt{>}, \texttt{<}) or outgoing (\texttt{/},\texttt{\char`\\}) arc from or to a specific direction (left, right) and $r_i$ is the incoming arc label of $w_i$. When using only the symbols in $B$, the bracketing encoding only covers sets of arcs with no crossing arcs in the same direction. \citet{strzyz-etal-2019-viable} extended $B$ with $B^*=\{\textcolor{BrickRed}{\texttt{/*}}, \textcolor{BrickRed}{\texttt{>*}}, \textcolor{BrickRed}{\texttt{<*}}, \textcolor{BrickRed}{\texttt{\char`\\*}}\}$ to only encode arcs of the second plane, thus supporting $k$-planar dependency trees where $k\leq 2$. 

\paragraph{4-bit and 7-bit encodings} \citet{gomez-rodriguez-etal-2023-4} proposed two bit-based encodings for projective and 2-planar dependency trees. In both variants, each label consists of two components: $\ell_i=(b_i, r_i)$, where $b_i\in\{0,1\}^m$ is a sequence of $m$ bits ($m=4$ or $m=7$, respectively) and $r_i$ is the dependency label of the incoming arc to $w_i$.

Let $b_i=(b_i^0, b_i^1, b_i^2, b_i^3)$ be the bit symbols of $\ell_i$ in the 4-bit encoding: $b_i^0$ is activated if $w_i$ has a left parent (otherwise, it is set to 0); $b_i^1$ is activated if $w_i$ is the outermost dependent of its parent in the same direction; $b_i^2$ is activated if $w_i$ has left dependents; and $b_i^3$ is activated if $w_i$ has right dependents. Figure \ref{fig:projective-tree-example} shows an example of the 4-bit encoding. Note that the label $\ell_5=(0100, \text{case})$ since the head of $w_5$ is located to the right ($7\to 5$) and $w_5$ is the leftmost dependent of $w_7$, and there are no arcs where $w_5$ is the head.

The 7-bit encoding extends the number of bits of the 4-bit encoding to 7 bits, so $b_i=(b_i^0,b_i^1, b_i^2, b_i^3, b_i^4, b_i^5, b_i^6)$. The first two bits $b_i^0b_i^1$ encode the plane and position of the head of $w_i$, so $b_i^0b_i^1\in\{00, 01, 10, 11\}$ if $w_i$ has a right or left head in the first plane ($b_i^0b_i^1=00$ or $b_i^0b_i^1=10$, respectively) or second plane ($b_i^0b_i^1=01$ or $b_i^0b_i^1=11$, respectively);  $b_i^2$ is activated if $w_i$ is the outermost dependent of its head in the same direction; $b_i^3$ and $b_i^4$ are activated if $w_i$ has left or right dependents in the first plane, respectively; and $b_i^5$ and $b_i^6$ are similarly activated for the dependencies of the second plane. See Figure \ref{fig:2-planar-tree-example} for an example of the 7-bit encoding in a 2-planar tree.

\subsection{Graph parsing}\label{appendix-graph-parsing}
Graph parsing relaxes the connectivity, acyclicity, and single-head constraints of a dependency tree. Given a dependency graph $G=(W,A)$, where $A=\{(h\overset{r}{\to}d : d \neq h, r \in\mathcal{R}\}$, we relied on the encodings proposed by \citet{ezquerro-etal-2024-dependency} to represent the arc information as a sequence of $n$ labels, where each label is always defined by two components $\ell_i=(x_i, \rho_i)$, where $x_i$ is configured depending on the encoding algorithm and $\rho_i$ remains constant as the concatenation of incoming arc labels ordered by the absolute position of the heads.

\paragraph{Relative and bracketing encoding} The relative and bracketing encodings for dependency trees \cite{strzyz-etal-2019-viable} can be directly applied to graphs, as displayed in Figure \ref{fig:simple-graph-example}. Due to the single-head constraint of dependency trees, under relative encoding each label only contains one head position. For graphs, since each node might have an arbitrary number of heads, the symbol $b_i$ is defined as the sorted sequence of head positions for $w_i$. The bracketing encoding is independent of tree constraints, as it only encodes the arc information for each individual node.

\paragraph{$\boldsymbol{4k}$ and $\boldsymbol{6k}$-bit encodings} The bit-based encodings proposed by \citet{ezquerro-etal-2024-dependency} encode a set of arcs $A$ by (i) first  distributing the arcs of $A$ into $k$ mutually-exclusive subsets $\{P_1 \cdots P_k\}$ where each $P_j\subseteq A$ satisfies certain conditions, (ii) then encoding each subset $P_j$ with $n$ symbols $(b_{1,j}\cdots b_{n,j})$ where each symbol $b_{i,j}\in\{0,1\}^{4|6}$ is a sequence of 4 or 6 bits, respectively; and (iii) finally concatenating each symbol at token level to produce $\ell_i=(b_{i,1}\cdots b_{i,k}, r_i)$.

In the $4k$-bit encoding, each subset $P_j$ satisfies that: (i) no arc in $P_j$ crosses another arc in $P_j$ in the same direction, and (ii) there is one and only one incoming arc per node. Then, the bit values of $b_{i,j}$ are assigned as in the 4-bit dependency encoding of \citet{gomez-rodriguez-etal-2023-4}. Since the second constraint cannot be satisfied if there are nodes without heads, the $4k$-bit encoding creates artificial arcs---connected to the previous node---that are labeled with a null type and discarded in the postprocessing step.

In the $6k$-bit encoding, each subset $P_j$ is instead constrained by (i) not having crossing arcs in the same direction, and (ii) each node having at most one incoming arc per direction. The symbol $b_{i,j}=(b_{i,j}^0b_{i,j}^1b_{i,j}^2b_{i,j}^3b_{i,j}^4b_{i,j}^5)$ activates the first (second) bit $b_{i,j}^0$ ($b_{i,j}^1$) with the presence of a left (right) parent of $w_i$ in $P_j$. The third (fourth) bit $b_{i,j}^2$ ($b_{i,j}^3$) is activated if $w_i$ is the farthest dependent of its left (right) head. The fifth (sixth) bit $b_{i,j}^4$ is activated if $w_i$ has left (right) dependents.

For illustrative examples and details regarding the bit-based graph encodings, we recommend reading \citet{ezquerro-etal-2024-dependency}.

\begin{figure}[h!]\small\centering
    \begin{dependency}[text only label, arc edge ]
        \begin{deptext}[column sep=5pt]
            $w_0$ \& $w_1$ \& $w_2$ \& $w_3$ \& $w_4$ \& $w_5$ \\[1em]
            \textbf{R}: \& - \& - \& (-3) \& (-3,1) \& (-3) \\[0.7em] 
            \textbf{B}: \& \texttt{\textcolor{BrickRed}{/*}<} \& \texttt{\textcolor{blue}{/**}} \& \texttt{\char`\\>} \& \texttt{\textcolor{BrickRed}{>*}<} \& \texttt{\char`\\} \texttt{\textcolor{blue}{>**}}\\
        \end{deptext}
        \depedge{1}{4}{}
        \depedge{4}{2}{}
        \depedge[edge style={BrickRed}, edge horizontal padding=-1ex]{2}{5}{}
        \depedge{6}{5}{}
        \depedge[edge style={blue}]{3}{6}{}
    \end{dependency}
    \caption{\label{fig:simple-graph-example}Dependency graph example encoded with the relative (\textbf{R}) and bracketing (\textbf{B}) encoding with $k=3$.}
\end{figure}

\section{Additional results}
In this section we present the detailed performance of our taggers on the five NLP tasks introduced in Section \ref{sec:experiments} (PoS tagging, NER, dependency parsing, constituency parsing and graph parsing). Tables \ref{tab:ptb} to \ref{tab:spmrl-sv} break down the constituency and PoS tagging performance of the original PTB (Table \ref{tab:ptb}) and CTB (Table \ref{tab:ctb}) annotations and the SPMRL datasets (Tables \ref{tab:spmrl-de}-\ref{tab:spmrl-sv}). Tables \ref{tab:ptb-dep} to \ref{tab:swedish-talbanken} show the dependency and PoS tagging performance of the PTB and CTB dependency conversions and the UD treebanks. Tables \ref{tab:en-dm} to \ref{tab:iwpt-2021-ta-ttb} show the graph parsing performance on the SDP \cite{oepen-etal-2015-semeval} and IWPT \cite{bouma-etal-2021-raw} selected datasets.

\begin{table}[h]\centering
    \fontsize{8pt}{8pt}\selectfont
    \setlength{\tabcolsep}{2pt}
    \renewcommand{\arraystretch}{1.1}
\begin{tabular}{|c|cccc|cccc|c|}
        \cline{2-10}
        \multicolumn{1}{c|}{}& \multicolumn{4}{c|}{\textbf{R}} & \multicolumn{4}{c|}{\textbf{T}} &\multirow{2}{*}{\textbf{POS}}\\
        \cline{2-9}
        \multicolumn{1}{c|}{}& UF & LF & UM & LM & UF & LF & UM & LM & \\
        \hline
        {\scriptsize $\rightarrow$} & 62.74 & 57.14 & 1.74 & 1.12 & 59.98 & 52.65 & 3.56 & 2.69 & 95.97 \\
        {\scriptsize $\overset{\text{\xlstm}}{\rightarrow}$} & 64.62 & 59.36 & 1.66 & 0.95 & 61.37 & 54.16 & 3.81 & 2.57 & 95.74 \\
        {\scriptsize $\leftrightarrow$} & 87.19 & 84.43 & 20.74 & 18.67 & 85.42 & 82.14 & 29.06 & 25.79 & 96.27 \\
        {\scriptsize $\overset{\text{\xlstm}}{\leftrightarrow}$} & 90.11 & 87.94 & 28.44 & 25.91 & 87.7 & 84.87 & 34.56 & 31.33 & 96.92 \\
        {\scriptsize DiT} & 94.47 & 92.76 & 40.98 & 36.3 & 93.37 & 90.03 & 50.91 & 39.98 & 95.83 \\
        {\scriptsize GaT} & \underline{\textbf{95.89}} & \underline{\textbf{94.96}} & \textbf{50.95} & \textbf{47.72} & \textbf{94.76} & \textbf{93.43} & \textbf{57.45} & \textbf{53.56} & \textbf{97.91} \\
        {\scriptsize SSD} & 68.25 & 63.16 & 2.32 & 1.28 & 65.08 & 58.72 & 4.64 & 3.23 & 97.33 \\
        \hline 
        {\scriptsize Base} & 95.84 & 94.88 & \underline{52.07} & \underline{49.21} & \underline{95.19} & \underline{93.97} & \underline{59.15} & \underline{55.5} & \underline{97.97} \\
        \hline 
    \end{tabular}
    \caption{\label{tab:ptb}PTB performance. We refer to the LSTM as ($\rightarrow$) and the BiLSTM as ($\leftrightarrow$). The symbol \xlstm\ indicates that the recurrent cell is replaced by the xLSTM. Same notation as in Table \ref{tab:con-results}.}
\end{table}
\begin{table}[h]\centering
    \fontsize{8pt}{8pt}\selectfont
    \setlength{\tabcolsep}{2pt}
    \renewcommand{\arraystretch}{1.1}
    \begin{tabular}{|c|cccc|cccc|c|}
        \cline{2-10}
        \multicolumn{1}{c|}{}& \multicolumn{4}{c|}{\textbf{R}} & \multicolumn{4}{c|}{\textbf{T}} &\multirow{2}{*}{\textbf{POS}}\\
        \cline{2-9}
        \multicolumn{1}{c|}{}& UF & LF & UM & LM & UF & LF & UM & LM & \\
        \hline
        {\scriptsize $\rightarrow$} & 68.11 & 54.79 & 9.16 & 7.43 & 65.47 & 55.46 & 9.58 & 8.85 & 92.71 \\
        {\scriptsize $\overset{\text{\xlstm}}{\rightarrow}$} & 68.99 & 55.63 & 8.64 & 6.54 & 66.62 & 56.33 & 9.21 & 8.85 & 91.95 \\
        {\scriptsize $\leftrightarrow$} & 83.48 & 76.32 & 14.45 & 13.09 & 81.29 & 74.91 & 15.03 & 13.46 & 92.74 \\
        {\scriptsize $\overset{\text{\xlstm}}{\leftrightarrow}$} & 86.32 & 79.64 & 18.85 & 16.54 & 84.13 & 78.8 & 19.84 & 17.75 & 93.55 \\
        {\scriptsize DiT} & 92.81 & 87.84 & 30.68 & 25.03 & 92.84 & 87.87 & 32.36 & 24.14 & 97.1 \\
        {\scriptsize GaT} & \underline{\textbf{94.16}} & \textbf{90.77} & \underline{\textbf{36.28}} & \underline{\textbf{32.41}} & \underline{\textbf{93.41}} & \underline{\textbf{93.52}} & \underline{\textbf{39.01}} & \underline{\textbf{35.08}} & \underline{\textbf{97.96}} \\
        {\scriptsize SSD} & 70.57 & 57.35 & 8.59 & 6.28 & 67.81 & 57.7 & 8.69 & 7.96 & 93.38 \\
        \hline 
        {\scriptsize Base} & 94.01 & \underline{90.84} & 36.07 & 32.25 & 92.81 & 90.04 & 36.39 & 32.36 & 97.79 \\
        \hline 
    \end{tabular}
    \caption{\label{tab:ctb}CTB performance. Notation as in Table \ref{tab:ptb}.}
\end{table}

\begin{table}[h]\centering
    \fontsize{8pt}{8pt}\selectfont
    \setlength{\tabcolsep}{2pt}
    \renewcommand{\arraystretch}{1.1}
    \begin{tabular}{|c|cccc|cccc|c|}
        \cline{2-10}
        \multicolumn{1}{c|}{}& \multicolumn{4}{c|}{\textbf{R}} & \multicolumn{4}{c|}{\textbf{T}} &\multirow{2}{*}{\textbf{POS}}\\
        \cline{2-9}
        \multicolumn{1}{c|}{}& UF & LF & UM & LM & UF & LF & UM & LM & \\
        \hline
        {\scriptsize $\rightarrow$} & 39.44 & 32.78 & 7.0 & 5.5 & 58.31 & 39.58 & 8.78 & 5.54 & 63.77 \\
        {\scriptsize $\overset{\text{\xlstm}}{\rightarrow}$} & 44.25 & 37.59 & 7.72 & 5.84 & 57.99 & 40.36 & 8.28 & 5.48 & 68.84 \\
        {\scriptsize $\leftrightarrow$} &  68.99 & 63.08 & 23.02 & 20.78 & 79.03 & 72.22 & 24.94 & 22.38 & 65.82 \\
        {\scriptsize $\overset{\text{\xlstm}}{\leftrightarrow}$} &75.4 & 70.68 & 30.08 & 27.72 & 83.13 & 77.29 & 32.92 & 29.64  & 71.48 \\
        {\scriptsize DiT} & 94.2 & 88.05 & 60.02 & 50.1 & 43.51 & 36.55 & 0.0 & 0.0   & 65.22 \\
        {\scriptsize GaT} &  \underline{\textbf{94.75}} & \underline{\textbf{91.22}} & \underline{\textbf{62.34}} & \underline{\textbf{58.14}} & \textbf{94.02} & \textbf{91.26} & \textbf{59.96} & \textbf{56.4} & \underline{\textbf{79.54}} \\
        {\scriptsize SSD} & 47.76 & 40.49 & 8.54 & 6.3 & 45.22 & 39.41 & 7.6 & 6.1 & 65.3 \\
        \hline 
        {\scriptsize Base} & 91.89 & 90.22 & 58.82 & 56.32 & \underline{94.12} & \underline{92.12} & \underline{60.24} & \underline{57.14} & 78.88 \\
        \hline 
    \end{tabular}
    \caption{\label{tab:spmrl-de}German-SPMRL performance. Same notation as in Table \ref{tab:ptb}.}
\end{table}

\begin{table}[h!]\centering
    \fontsize{8pt}{8pt}\selectfont
    \setlength{\tabcolsep}{2pt}
    \renewcommand{\arraystretch}{1.1}
    \begin{tabular}{|c|cccc|cccc|c|}
        \cline{2-10}
        \multicolumn{1}{c|}{}& \multicolumn{4}{c|}{\textbf{R}} & \multicolumn{4}{c|}{\textbf{T}} &\multirow{2}{*}{\textbf{POS}}\\
        \cline{2-9}
        \multicolumn{1}{c|}{}& UF & LF & UM & LM & UF & LF & UM & LM & \\
        \hline
        {\scriptsize $\rightarrow$} & 65.76 & 52.21 & 0.32 & 0.11 & 66.56 & 46.64 & 0.42 & 0.11& 67.28 \\
        {\scriptsize $\overset{\text{\xlstm}}{\rightarrow}$} & 65.77 & 50.9 & 0.74 & 0.21 & 65.91 & 55.81 & 1.06 & 0.63  & 70.26 \\
        {\scriptsize $\leftrightarrow$} & 77.29 & 66.19 & 7.19 & 4.76 & 78.98 & 72.58 & 11.63 & 8.77 & 32.72 \\
        {\scriptsize $\overset{\text{\xlstm}}{\leftrightarrow}$} &  79.64 & 68.78 & 10.78 & 5.92 & 81.33 & 74.62 & 15.75 & 11.73 & 71.21  \\
        {\scriptsize DiT} &  88.07 & 80.2 & 19.87 & 10.25 & 87.03 & 79.16 & 18.67 & 8.54  & 54.07 \\
        {\scriptsize GaT} & \textbf{92.21} & \underline{\textbf{89.24}} & \textbf{37.84} & \textbf{31.92} & \textbf{89.58} & \textbf{85.55} & \textbf{25.26} & \textbf{19.87} & \textbf{72.99} \\
        {\scriptsize SSD} & 63.23 & 48.55 & 0.32 & 0.0 & 55.81 & 37.04 & 0.63 & 0.32  & 59.47 \\
        \hline
        {\scriptsize Base} &  \underline{92.28} & 89.17 & \underline{40.38} & \underline{32.66} & \underline{92.4} & \underline{89.29} & \underline{47.78} & \underline{37.74} & \underline{74.14} \\
        \hline 
    \end{tabular}
    \caption{\label{tab:spmrl-eu}Basque-SPMRL performance. Same notation as in Table \ref{tab:ptb}.}
\end{table}
\begin{table}[h]\centering
    \fontsize{8pt}{8pt}\selectfont
    \setlength{\tabcolsep}{2pt}
    \renewcommand{\arraystretch}{1.1}
    \begin{tabular}{|c|cccc|cccc|c|}
        \cline{2-10}
        \multicolumn{1}{c|}{}& \multicolumn{4}{c|}{\textbf{R}} & \multicolumn{4}{c|}{\textbf{T}} &\multirow{2}{*}{\textbf{POS}}\\
        \cline{2-9}
        \multicolumn{1}{c|}{}& UF & LF & UM & LM & UF & LF & UM & LM & \\
        \hline
        {\scriptsize $\rightarrow$} &  53.67 & 43.05 & 1.34 & 0.98 & 16.46 & 1.67 & 0.2 & 0.08 & 79.03 \\
        {\scriptsize $\overset{\text{\xlstm}}{\rightarrow}$} & 54.12 & 43.75 & 1.5 & 0.87 & 56.87 & 46.66 & 3.66 & 2.4 & 80.12 \\
        {\scriptsize $\leftrightarrow$} & 72.6 & 65.86 & 10.47 & 9.45 & 72.32 & 66.37 & 9.88 & 8.62 & 79.36 \\
        {\scriptsize $\overset{\text{\xlstm}}{\leftrightarrow}$} &  76.6 & 70.79 & 12.48 & 11.02 & 77.61 & 72.63 & 13.73 & 12.0 & 85.55 \\
        {\scriptsize DiT} & 88.04 & 84.91 & 21.37 & 18.38 & 62.75 & 52.5 & 2.44 & 1.3  & 78.08 \\
        {\scriptsize GaT} &\underline{\textbf{88.82}} & \underline{\textbf{86.52}} & \textbf{21.96} & \textbf{20.15} & \textbf{86.13} & \textbf{83.25} & \textbf{21.21} & \textbf{19.36} & \underline{\textbf{88.6}} \\
        {\scriptsize SSD} & 57.35 & 47.44 & 1.73 & 1.34 & 52.46 & 44.23 & 1.77 & 1.18  & 80.05 \\
        \hline 
        {\scriptsize Base} &  87.23 & 83.39 & \underline{22.98} & \underline{20.78} & \underline{87.74} & \underline{84.79} & \underline{26.01} & \underline{24.12}  & 88.55 \\
        \hline 
    \end{tabular}
    \caption{\label{tab:spmrl-fr}French-SPMRL performance. Same notation as in Table \ref{tab:ptb}.}
\end{table}
\begin{table}[h]\centering
    \fontsize{8pt}{8pt}\selectfont
    \setlength{\tabcolsep}{2pt}
    \renewcommand{\arraystretch}{1.1}
    \begin{tabular}{|c|cccc|cccc|c|}
        \cline{2-10}
        \multicolumn{1}{c|}{}& \multicolumn{4}{c|}{\textbf{R}} & \multicolumn{4}{c|}{\textbf{T}} &\multirow{2}{*}{\textbf{POS}}\\
        \cline{2-9}
        \multicolumn{1}{c|}{}& UF & LF & UM & LM & UF & LF & UM & LM & \\
        \hline
        {\scriptsize $\rightarrow$} & 82.46 & 71.6 & 1.68 & 0.42 & 83.36 & 79.53 & 3.21 & 1.82& 84.43 \\
        {\scriptsize $\overset{\text{\xlstm}}{\rightarrow}$} & 82.02 & 70.06 & 1.82 & 0.42 & 82.97 & 78.62 & 3.35 & 1.4 & 81.99 \\
        {\scriptsize $\leftrightarrow$} & 6.88 & 75.38 & 3.77 & 2.09 & 78.6 & 78.0 & 0.7 & 0.42 & 84.54 \\
        {\scriptsize $\overset{\text{\xlstm}}{\leftrightarrow}$} & 87.54 & 75.91 & 6.15 & 3.35 & 88.39 & 85.25 & 9.92 & 7.96 & 84.01 \\
        {\scriptsize DiT} & 94.54 & 91.17 & 18.44 & 10.47 & 90.42 & 82.78 & 11.45 & 3.77 & 90.98 \\
        {\scriptsize GaT} &  \underline{\textbf{95.15}} & \underline{\textbf{92.47}} & \underline{\textbf{21.51}} & \underline{\textbf{16.34}} & \underline{\textbf{94.02}} & \underline{\textbf{91.1}} & \underline{\textbf{26.4}} & \underline{\textbf{19.83}} & \textbf{95.36} \\
        {\scriptsize SSD} & 79.22 & 61.61 & 1.12 & 0.14 & 77.72 & 58.83 & 0.7 & 0.28 & 69.11 \\
        \hline
        {\scriptsize Base} & 95.09 & 92.43 & 19.55 & 14.39 & 93.85 & 90.82 & 24.86 & 18.3 & \underline{95.36} \\
        \hline 
    \end{tabular}
    \caption{\label{tab:spmrl-he}Hebrew-SPMRL performance. Same notation as in Table \ref{tab:ptb}.}
\end{table}
\begin{table}[h]\centering
    \fontsize{8pt}{8pt}\selectfont
    \setlength{\tabcolsep}{2pt}
    \renewcommand{\arraystretch}{1.1}
    \begin{tabular}{|c|cccc|cccc|c|}
        \cline{2-10}
        \multicolumn{1}{c|}{}& \multicolumn{4}{c|}{\textbf{R}} & \multicolumn{4}{c|}{\textbf{T}} &\multirow{2}{*}{\textbf{POS}}\\
        \cline{2-9}
        \multicolumn{1}{c|}{}& UF & LF & UM & LM & UF & LF & UM & LM & \\
        \hline
        {\scriptsize $\rightarrow$} & 73.66 & 60.32 & 4.26 & 2.18 & 74.09 & 60.19 & 3.96 & 1.78 & 15.34 \\
        {\scriptsize $\overset{\text{\xlstm}}{\rightarrow}$} & 71.24 & 57.42 & 2.87 & 1.09 & 72.73 & 59.06 & 2.28 & 0.89  & \textbf{75.63} \\
        {\scriptsize $\leftrightarrow$} & 9.65 & 8.78 & 0.1 & 0.0 & 79.89 & 70.23 & 11.6 & 7.23 & 58.15 \\
        {\scriptsize $\overset{\text{\xlstm}}{\leftrightarrow}$} & 80.06 & 70.7 & 11.2 & 7.93 & 80.25 & 71.83 & 12.78 & 8.92 & 71.05 \\
        {\scriptsize DiT} &  94.41 & 89.4 & 45.79 & 33.0 & 90.14 & 79.83 & 15.98 & 3.71 & 61.17 \\
        {\scriptsize GaT} &  \underline{\textbf{95.11}} & \textbf{92.42} & \textbf{51.64} & \underline{\textbf{43.11}} & \underline{\textbf{93.29}} & \textbf{90.45} & \textbf{35.68} & \textbf{29.93} & 74.28 \\
        {\scriptsize SSD} &  70.8 & 55.78 & 2.38 & 0.99 & 65.5 & 49.56 & 1.39 & 0.1 & 60.05 \\
        \hline 
        {\scriptsize Base} & 95.1 & \underline{92.47} & \underline{52.82} & 43.11 & 93.28 & \underline{90.97} & \underline{39.84} & \underline{33.99}& \underline{75.71} \\
        \hline 
    \end{tabular}
    \caption{\label{tab:spmrl-hu}Hungarian-SPMRL performance. Same notation as in Table \ref{tab:ptb}.}
\end{table}

\begin{table}[h]\centering
    \fontsize{8pt}{8pt}\selectfont
    \setlength{\tabcolsep}{2pt}
    \renewcommand{\arraystretch}{1.1}
    \begin{tabular}{|c|cccc|cccc|c|}
        \cline{2-10}
        \multicolumn{1}{c|}{}& \multicolumn{4}{c|}{\textbf{R}} & \multicolumn{4}{c|}{\textbf{H}} &\multirow{2}{*}{\textbf{POS}}\\
        \cline{2-9}
        \multicolumn{1}{c|}{}& UF & LF & UM & LM & UF & LF & UM & LM & \\
        \hline
        {\scriptsize $\rightarrow$} & 60.69 & 52.52 & 0.83 & 0.31 & 66.14 & 54.81 & 4.5 & 1.57 & 60.51 \\
        {\scriptsize $\overset{\text{\xlstm}}{\rightarrow}$} & 58.93 & 50.47 & 0.7 & 0.26 & 66.75 & 56.32 & 4.81 & 2.01  & 65.24 \\
        {\scriptsize $\leftrightarrow$} & 67.23 & 60.2 & 8.48 & 5.55 & 74.68 & 67.72 & 12.42 & 8.92  & 49.74 \\
        {\scriptsize $\overset{\text{\xlstm}}{\leftrightarrow}$} & 68.06 & 61.61 & 9.49 & 6.95 & 76.66 & 71.0 & 16.0 & 12.51  & 65.52 \\
        {\scriptsize DiT} & 89.29 & 85.18 & 34.89 & 27.2 & 85.66 & 79.16 & 29.09 & 17.91 & 46.65 \\
        {\scriptsize GaT} & \underline{\textbf{89.95}} & \underline{\textbf{87.6}} & \underline{\textbf{40.93}} & \underline{\textbf{35.33}} & \textbf{87.15} & \textbf{84.01} & \textbf{31.31} & \textbf{25.27}  & \underline{\textbf{66.56}} \\
        {\scriptsize SSD} & 52.49 & 41.18 & 0.0 & 0.0 & 47.74 & 31.46 & 0.13 & 0.0 & 33.29 \\
        \hline 
        {\scriptsize Base} &89.43 & 86.92 & 38.61 & 33.36 & \underline{89.77} & \underline{86.93} & \underline{40.27} & \underline{33.54} & 66.42 \\
        \hline 
    \end{tabular}
    \caption{\label{tab:spmrl-ko}Korean-SPMRL performance. Same notation as in Table \ref{tab:ptb}.}
\end{table}

\begin{table}[h]\centering    
    \fontsize{8pt}{8pt}\selectfont
    \setlength{\tabcolsep}{2pt}
    \renewcommand{\arraystretch}{1.1}
    \begin{tabular}{|c|cccc|cccc|c|}
        \cline{2-10}
        \multicolumn{1}{c|}{}& \multicolumn{4}{c|}{\textbf{R}} & \multicolumn{4}{c|}{\textbf{T}} &\multirow{2}{*}{\textbf{POS}}\\
        \cline{2-9}
        \multicolumn{1}{c|}{}& UF & LF & UM & LM & UF & LF & UM & LM & \\
        \hline
        {\scriptsize $\rightarrow$}  &  80.46 & 63.39 & 3.28 & 1.09 & 77.85 & 64.95 & 5.11 & 1.82  & 61.76 \\
        {\scriptsize $\overset{\text{\xlstm}}{\rightarrow}$} 
        & 79.75 & 59.61 & 3.89 & 1.7 & 82.4 & 73.19 & 9.49 & 4.74 & 64.7 \\
        {\scriptsize $\leftrightarrow$} & 72.62 & 30.46 & 0.0 & 0.0 & 91.24 & 87.65 & 32.73 & 28.1 & 42.71 \\
        {\scriptsize $\overset{\text{\xlstm}}{\leftrightarrow}$}& 88.58 & 74.6 & 26.4 & 14.48 & 92.39 & 89.47 & 38.32 & 34.06 & 64.34 \\
        {\scriptsize DiT} & \underline{\textbf{97.04}} & 93.43 & 59.61 & 48.42 & \textbf{97.03} & \textbf{94.07} & \textbf{62.9} & \textbf{48.3} & 48.55 \\
        {\scriptsize GaT}  & 96.9 & \textbf{94.39} & \textbf{59.85} & \textbf{49.15} & 96.28 & 93.6 & 48.66 & 37.1  & \textbf{66.1} \\
        {\scriptsize SSD} &  79.5 & 62.77 & 1.09 & 0.36 & 75.2 & 53.23 & 2.55 & 1.22 & 53.52 \\
        \hline 
        {\scriptsize Base} & 96.9 & \underline{94.46} & \underline{60.95} & \underline{51.46} & \underline{97.49} & \underline{95.85} & \underline{69.46} & \underline{62.04}  & \underline{66.92} \\
        \hline 
    \end{tabular}
    \caption{\label{tab:spmrl-pl}Polish-SPMRL performance. Same notation as in Table \ref{tab:ptb}.}
\end{table}

\begin{table}[h!]\centering\footnotesize        
    \setlength{\tabcolsep}{2pt}
    \renewcommand{\arraystretch}{1.1}
    \begin{tabular}{|c|cccc|cccc|c|}
        \cline{2-10}
        \multicolumn{1}{c|}{}& \multicolumn{4}{c|}{\textbf{R}} & \multicolumn{4}{c|}{\textbf{T}} &\multirow{2}{*}{\textbf{POS}}\\
        \cline{2-9}
        \multicolumn{1}{c|}{}& UF & LF & UM & LM & UF & LF & UM & LM & \\
        \hline
        {\scriptsize $\rightarrow$} & 55.06 & 48.81 & 5.11 & 3.9 & 59.28 & 46.26 & 6.01 & 4.35  & 80.51 \\
        {\scriptsize $\overset{\text{\xlstm}}{\rightarrow}$} & 54.23 & 47.43 & 3.75 & 2.7 & 59.33 & 46.31 & 6.46 & 4.65  & 76.46 \\
        {\scriptsize $\leftrightarrow$} &  13.7 & 12.52 & 0.15 & 0.0 & 24.87 & 14.94 & 3.75 & 0.15  & 78.39 \\
        {\scriptsize $\overset{\text{\xlstm}}{\leftrightarrow}$} & 70.88 & 63.03 & 12.76 & 6.01 & 73.46 & 62.72 & 13.21 & 10.36 & 80.08 \\
        {\scriptsize DiT} &90.46 & 81.26 & 34.08 & 21.92 & 84.59 & 66.95 & 19.84 & 7.49 & 89.82 \\
        {\scriptsize GaT} &\underline{\textbf{90.77}} & \textbf{88.02} & \textbf{39.19} & \textbf{34.53} & \textbf{88.31} & \textbf{84.53} & \textbf{42.19} & \textbf{37.84} & \textbf{94.38} \\
        {\scriptsize SSD} & 51.51 & 44.68 & 3.3 & 2.55 & 51.56 & 40.57 & 3.45 & 1.95 & 77.36 \\
        \hline 
        {\scriptsize Base} & 90.67 & \underline{88.12} & \underline{41.14} & \underline{36.49} & \underline{89.69} & \underline{86.07} & \underline{45.35} & \underline{41.29} & \underline{94.47} \\
        \hline 
    \end{tabular}
    \caption{\label{tab:spmrl-sv}Swedish-SPMRL performance. Same notation as in Table \ref{tab:ptb}.}
\end{table}

\begin{table*}[h!]\footnotesize\centering
    \setlength{\tabcolsep}{2pt}
    \renewcommand{\arraystretch}{1.1}
    \begin{tabular}{|c|cccc|cccc|cccc|cccc|}
        \cline{2-17}
        \multicolumn{1}{c|}{} & \multicolumn{4}{c|}{\textbf{B}} & \multicolumn{4}{c|}{\textbf{B4}} & \multicolumn{4}{c|}{\textbf{B7}} & \multicolumn{4}{c|}{\textbf{H}}\\
        \cline{2-17}
        \multicolumn{1}{c|}{} & UAS & LAS & UM & LM & UAS & LAS & UM & LM& UAS & LAS & UM & LM& UAS & LAS & UM & LM \\
        \hline 
        {\scriptsize $\rightarrow$} & 65.97 & 60.01 & 6.17 & 3.56 & 66.28 & 60.42 & 6.0 & 3.1 & 65.99 & 60.05 & 5.46 & 2.86 & 58.86 & 53.96 & 5.75 & 3.02 \\
        {\scriptsize $\overset{\text{\xlstm}}{\rightarrow}$} & 67.64 & 61.53 & 6.66 & 3.68 & 68.35 & 62.63 & 6.33 & 2.86 & 68.28 & 62.65 & 6.58 & 3.52 & 61.71 & 57.05 & 5.75 & 2.94 \\
        {\scriptsize $\leftrightarrow$} & 90.16 & 88.51 & 35.97 & 30.67 & 89.51 & 87.92 & 33.94 & 29.01 & 90.05 & 88.51 & 36.34 & 31.37 & 88.22 & 86.35 & 32.49 & 26.82 \\
        {\scriptsize $\overset{\text{\xlstm}}{\leftrightarrow}$} & 91.92 & 90.41 & 41.35 & 35.68 & 91.51 & 90.03 & 40.73 & 34.98 & 91.34 & 90.01 & 39.94 & 35.43 & 91.45 & 90.14 & 42.09 & 36.67 \\
        {\scriptsize DiT} & 89.39 & 86.83 & 35.02 & 26.2 & 94.4 & 92.31 & 55.05 & 40.31 & 94.79 & 92.39 & 57.66 & 41.39 & 94.44 & 92.04 & 55.26 & 38.37 \\
        {\scriptsize GaT} & \underline{\textbf{95.77}} & \underline{\textbf{94.13}} & \underline{\textbf{61.96}} & \underline{\textbf{49.05}} & \textbf{95.87} & \textbf{94.19} & \underline{\textbf{63.37}} & \underline{\textbf{49.92}} & \underline{\textbf{95.85}} & \underline{\textbf{94.19}} & \underline{\textbf{63.87}} & \underline{\textbf{50.37}} & \underline{\textbf{95.78}} & \underline{\textbf{94.13}} & \underline{\textbf{63.78}} & \underline{\textbf{49.75}} \\
\hline 
    \end{tabular}
    \caption{\label{tab:ptb-dep}Performance in the PTB with dependency annotations. Same notation as in Tables \ref{tab:pos-results}, \ref{tab:dep-results} and \ref{tab:ptb}.}
\end{table*}

\begin{table*}[h]\footnotesize\centering
    \setlength{\tabcolsep}{2pt}
    \renewcommand{\arraystretch}{1.1}
    \begin{tabular}{|c|cccc|cccc|cccc|cccc|}
        \cline{2-17}
        \multicolumn{1}{c|}{} & \multicolumn{4}{c|}{\textbf{B}} & \multicolumn{4}{c|}{\textbf{B4}} & \multicolumn{4}{c|}{\textbf{B7}} & \multicolumn{4}{c|}{\textbf{H}}\\
        \cline{2-17}
        \multicolumn{1}{c|}{} & UAS & LAS & UM & LM & UAS & LAS & UM & LM& UAS & LAS & UM & LM& UAS & LAS & UM & LM \\
        \hline 
        {\scriptsize $\rightarrow$} & 57.75 & 45.13 & 9.97 & 7.19 & 58.13 & 45.42 & 10.71 & 7.19 & 58.13 & 45.42 & 10.71 & 7.19 & 54.96 & 42.52 & 11.29 & 6.88 \\
        {\scriptsize $\overset{\text{\xlstm}}{\rightarrow}$} & 61.81 & 50.41 & 12.44 & 8.87 & 61.12 & 49.5 & 12.86 & 9.08 & 61.12 & 49.5 & 12.86 & 9.08 & 56.65 & 45.36 & 12.34 & 8.82 \\
        {\scriptsize $\leftrightarrow$} & 82.12 & 79.3 & 25.93 & 21.21 & 81.9 & 79.3 & 26.93 & 22.52 & 81.9 & 79.3 & 26.93 & 22.52 & 79.42 & 76.31 & 25.14 & 20.47 \\
        {\scriptsize $\overset{\text{\xlstm}}{\leftrightarrow}$} & 84.16 & 82.18 & 30.08 & 25.3 & 83.19 & 80.82 & 27.82 & 23.57 & 83.19 & 80.82 & 27.82 & 23.57 & 82.15 & 79.68 & 29.29 & 24.36 \\
        {\scriptsize DiT} & 84.29 & 82.21 & 28.29 & 24.51 & 90.17 & 87.93 & 42.2 & 33.91 & 89.63 & 87.42 & 40.79 & 33.23 & 89.28 & 86.65 & 41.78 & 32.34 \\
        {\scriptsize GaT} & \underline{\textbf{90.86}} & \underline{\textbf{88.84}} & \underline{\textbf{42.05}} & \underline{\textbf{34.44}} & \underline{\textbf{91.17}} & \underline{\textbf{89.19}} & \underline{\textbf{46.14}} & \underline{\textbf{37.59}} & \textbf{90.27} & \textbf{88.14} & \textbf{43.25} & \underline{\textbf{35.17}} & \underline{\textbf{90.83}} & \underline{\textbf{88.57}} & \underline{\textbf{45.83}} & \underline{\textbf{36.17}} \\
\hline 
    \end{tabular}
    \caption{\label{tab:ctb-dep}Performance in the CTB with dependency annotations. Same notation as in Table \ref{tab:ptb-dep}.}
\end{table*}
\begin{table*}[h]\footnotesize\centering
    \setlength{\tabcolsep}{2pt}
    \renewcommand{\arraystretch}{1.1}
    \begin{tabular}{|c|cccc|cccc|cccc|cccc|cc|}
        \cline{2-19}
        \multicolumn{1}{c|}{} & \multicolumn{4}{c|}{\textbf{B}} & \multicolumn{4}{c|}{\textbf{B4}} & \multicolumn{4}{c|}{\textbf{B7}} & \multicolumn{4}{c|}{\textbf{H}} & \multirow{2}{*}{U} & \multirow{2}{*}{\textbf{X}}\\ 
        \cline{2-17}
        \multicolumn{1}{c|}{} & UAS & LAS & UM & LM & UAS & LAS & UM & LM& UAS & LAS & UM & LM& UAS & LAS & UM & LM & &\\
        \hline 
        {\scriptsize $\rightarrow$} & 60.99 & 51.89 & 7.57 & 2.46 & 61.49 & 53.4 & 7.37 & 2.76 & 60.94 & 52.63 & 6.96 & 2.46 & 55.91 & 48.96 & 8.5 & 3.48 & 90.31 & 89.99 \\
        {\scriptsize $\overset{\text{\xlstm}}{\rightarrow}$} & 69.43 & 60.81 & 11.87 & 4.81 & 64.12 & 56.59 & 9.62 & 3.17 & 63.9 & 56.09 & 11.05 & 5.12 & 58.27 & 51.98 & 8.5 & 3.68 & 88.96 & 88.31 \\
        {\scriptsize $\leftrightarrow$} & 77.22 & 67.42 & 18.83 & 7.27 & 76.44 & 67.25 & 19.86 & 9.01 & 76.82 & 66.91 & 20.06 & 8.19 & 75.63 & 67.21 & 21.8 & 9.72 & 91.52 & 91.79 \\
        {\scriptsize $\overset{\text{\xlstm}}{\leftrightarrow}$} & 81.87 & 74.22 & 25.18 & 13.2 & 81.68 & 73.87 & 27.43 & 14.02 & 82.9 & 76.07 & 28.56 & 15.97 & 81.95 & 75.8 & 29.99 & 16.99 & 91.36 & 92.26 \\
        {\scriptsize DiT} & 75.74 & 70.28 & 20.78 & 11.77 & \underline{\textbf{86.44}} & 80.75 & \textbf{38.38} & 19.45 & 85.38 & 79.74 & 37.67 & 20.06 & 85.14 & 79.22 & 35.62 & 16.89 & \textbf{96.62} & 94.62 \\
        {\scriptsize GaT} & \underline{\textbf{87.67}} & \underline{\textbf{82.65}} & \underline{\textbf{39.3}} & \underline{\textbf{22.82}} & 86.36 & \underline{\textbf{81.67}} & 38.18 & \underline{\textbf{22.52}} & \underline{\textbf{87.92}} & \underline{\textbf{83.14}} & \underline{\textbf{42.99}} & \underline{\textbf{24.56}} & \underline{\textbf{87.55}} & \underline{\textbf{82.68}} & \underline{\textbf{42.78}} & \underline{\textbf{23.34}} & 96.21 & \textbf{97.42} \\
\hline 
    \end{tabular}
    \caption{\label{tab:german-gsd}Performance in the German-GSD. Same notation as in Table \ref{tab:ptb-dep}.}
\end{table*}
\begin{table*}[h]\footnotesize\centering
    \setlength{\tabcolsep}{2pt}
    \renewcommand{\arraystretch}{1.1}
    \begin{tabular}{|c|cccc|cccc|cccc|cccc|c|}
        \cline{2-18}
        \multicolumn{1}{c|}{} & \multicolumn{4}{c|}{\textbf{B}} & \multicolumn{4}{c|}{\textbf{B4}} & \multicolumn{4}{c|}{\textbf{B7}} & \multicolumn{4}{c|}{\textbf{H}} & \multirow{2}{*}{\textbf{U}}\\ 
        \cline{2-17}
        \multicolumn{1}{c|}{} & UAS & LAS & UM & LM & UAS & LAS & UM & LM& UAS & LAS & UM & LM& UAS & LAS & UM & LM & \\
        \hline 
        {\scriptsize $\rightarrow$} & 52.02 & 41.85 & 4.56 & 2.06 & 54.21 & 44.77 & 4.84 & 2.11 & 54.75 & 44.51 & 5.34 & 2.22 & 44.97 & 36.92 & 4.67 & 1.5 & 87.65 \\
        {\scriptsize $\overset{\text{\xlstm}}{\rightarrow}$} & 55.7 & 45.48 & 5.89 & 2.22 & 53.1 & 43.31 & 5.11 & 2.06 & 53.72 & 43.86 & 5.28 & 1.78 & 44.53 & 36.63 & 5.23 & 2.17 & 85.06 \\
        {\scriptsize $\leftrightarrow$} & 70.03 & 57.46 & 12.95 & 4.5 & 70.43 & 58.34 & 15.4 & 5.06 & 69.11 & 56.36 & 14.01 & 4.17 & 67.93 & 56.76 & 17.01 & 5.84 & 81.76 \\
        {\scriptsize $\overset{\text{\xlstm}}{\leftrightarrow}$} & 78.18 & 68.11 & 22.68 & 8.78 & 77.05 & 67.9 & 22.57 & 10.56 & 78.0 & 68.44 & 22.9 & 9.34 & 75.91 & 66.8 & 24.18 & 9.73 & 85.56 \\
        {\scriptsize DiT} & 59.0 & 53.77 & 4.17 & 2.33 & 84.69 & 79.43 & 42.3 & 24.51 & 83.63 & 78.86 & 40.08 & 25.57 & 85.25 & 78.47 & 41.47 & 19.46 & 93.54 \\
        {\scriptsize GaT} & \textbf{86.0} & \textbf{81.76} & \textbf{42.47} & \textbf{28.74} & \underline{\textbf{86.2}} & \underline{\textbf{81.93}} & \underline{\textbf{45.41}} & \underline{\textbf{31.13}} & \textbf{86.64} & \textbf{82.54} & \textbf{46.47} & \underline{\textbf{31.63}} & \textbf{86.06} & \textbf{81.76} & \textbf{44.91} & \textbf{29.41} & \underline{\textbf{95.54}} \\
\hline 
    \end{tabular}
    \caption{\label{tab:basque-bdt}Performance in the Basque-BDT. Same notation as in Table \ref{tab:ptb-dep}.}
\end{table*}

\begin{table*}[h]\footnotesize\centering
    \setlength{\tabcolsep}{2pt}
    \renewcommand{\arraystretch}{1.1}
    \begin{tabular}{|c|cccc|cccc|cccc|cccc|c|}
        \cline{2-18}
        \multicolumn{1}{c|}{} & \multicolumn{4}{c|}{\textbf{B}} & \multicolumn{4}{c|}{\textbf{B4}} & \multicolumn{4}{c|}{\textbf{B7}} & \multicolumn{4}{c|}{\textbf{H}} & \multirow{2}{*}{\textbf{U}}\\ 
        \cline{2-17}
        \multicolumn{1}{c|}{} & UAS & LAS & UM & LM & UAS & LAS & UM & LM& UAS & LAS & UM & LM& UAS & LAS & UM & LM & \\
        \hline 
        {\scriptsize $\rightarrow$} & 73.06 & 65.31 & 4.81 & 1.44 & 73.63 & 66.27 & 4.81 & 2.16 & 73.61 & 66.07 & 3.61 & 1.2 & 66.27 & 60.32 & 4.33 & 1.44 & 94.81 \\
        {\scriptsize $\overset{\text{\xlstm}}{\rightarrow}$} & 76.22 & 69.62 & 5.77 & 2.88 & 75.34 & 68.39 & 5.05 & 1.68 & 74.99 & 68.35 & 7.45 & 3.85 & 67.99 & 62.0 & 5.53 & 1.92 & 94.04 \\
        {\scriptsize $\leftrightarrow$} & 86.59 & 80.85 & 20.43 & 10.1 & 86.36 & 81.8 & 19.47 & 12.74 & 84.47 & 78.14 & 18.27 & 7.45 & 85.02 & 79.57 & 21.39 & 10.1 & 94.37 \\
        {\scriptsize $\overset{\text{\xlstm}}{\leftrightarrow}$} & 88.44 & 84.61 & 22.84 & 15.62 & 88.27 & 84.29 & 26.44 & 16.35 & 88.82 & 84.75 & 24.04 & 13.22 & 89.28 & 84.9 & 29.57 & 16.59 & 95.62 \\
        {\scriptsize DiT} & 88.95 & 85.94 & 32.93 & 23.08 & 93.47 & 90.57 & 46.39 & \textbf{32.21} & 93.12 & 90.34 & 43.03 & 30.05 & 94.32 & 90.78 & 49.04 & 29.81 & 98.09 \\
        {\scriptsize GaT} & \underline{\textbf{94.64}} & \underline{\textbf{91.99}} & \underline{\textbf{47.84}} & \underline{\textbf{33.89}} & \underline{\textbf{93.99}} & \textbf{91.05} & \textbf{49.04} & 32.21 & \underline{\textbf{94.47}} & \underline{\textbf{91.94}} & \underline{\textbf{50.0}} & \underline{\textbf{36.78}} & \underline{\textbf{94.84}} & \underline{\textbf{91.97}} & \textbf{50.0} & \underline{\textbf{35.82}} & \textbf{98.42} \\
\hline 
    \end{tabular}
    \caption{\label{tab:french-gsd}Performance in the French-GSD. Same notation as in Table \ref{tab:ptb-dep}.}
\end{table*}

\begin{table*}[h]\footnotesize\centering
    \setlength{\tabcolsep}{2pt}
    \renewcommand{\arraystretch}{1.1}
    \begin{tabular}{|c|cccc|cccc|cccc|cccc|c|}
        \cline{2-18}
        \multicolumn{1}{c|}{} & \multicolumn{4}{c|}{\textbf{B}} & \multicolumn{4}{c|}{\textbf{B4}} & \multicolumn{4}{c|}{\textbf{B7}} & \multicolumn{4}{c|}{\textbf{H}} & \multirow{2}{*}{\textbf{U}}\\ 
        \cline{2-17}
        \multicolumn{1}{c|}{} & UAS & LAS & UM & LM & UAS & LAS & UM & LM& UAS & LAS & UM & LM& UAS & LAS & UM & LM & \\
        \hline 
        {\scriptsize $\rightarrow$} & 68.43 & 63.89 & 2.65 & 1.43 & 68.67 & 64.25 & 1.83 & 0.81 & 69.49 & 64.59 & 1.43 & 1.22 & 59.31 & 55.73 & 0.81 & 0.61 & 92.47 \\
        {\scriptsize $\overset{\text{\xlstm}}{\rightarrow}$} & 70.95 & 66.75 & 2.24 & 1.02 & 69.77 & 65.84 & 1.63 & 1.02 & 70.11 & 65.71 & 2.04 & 0.81 & 62.16 & 58.56 & 1.43 & 0.81 & 90.3 \\
        {\scriptsize $\leftrightarrow$} & 77.55 & 71.56 & 8.35 & 4.07 & 80.04 & 74.02 & 11.61 & 5.7 & 78.76 & 72.38 & 9.16 & 4.07 & 74.87 & 68.54 & 11.0 & 4.28 & 91.41 \\
        {\scriptsize $\overset{\text{\xlstm}}{\leftrightarrow}$} & 82.07 & 78.25 & 13.44 & 7.74 & 83.35 & 79.3 & 16.29 & 9.37 & 84.11 & 80.31 & 18.13 & 12.02 & 81.69 & 77.58 & 15.07 & 9.57 & 92.05 \\
        {\scriptsize DiT} & 81.58 & 77.97 & 12.83 & 8.15 & 89.67 & 85.3 & 30.55 & 17.52 & 90.42 & 86.43 & 32.59 & 19.55 & 90.98 & 85.87 & 34.62 & 15.68 & 95.24 \\
        {\scriptsize GaT} & \textbf{91.75} & \textbf{88.43} & \underline{\textbf{38.09}} & \textbf{24.85} & \underline{\textbf{92.5}} & \underline{\textbf{89.55}} & \underline{\textbf{39.92}} & \underline{\textbf{27.29}} & \textbf{91.33} & \textbf{88.18} & \textbf{36.46} & \textbf{23.63} & \textbf{92.19} & \textbf{88.98} & \textbf{40.33} & \textbf{25.87} & \underline{\textbf{97.88}} \\
\hline 
    \end{tabular}
    \caption{\label{tab:hebrew-htb}Performance in the Hebrew-HTB. Same notation as in Table \ref{tab:ptb-dep}.}
\end{table*}

\begin{table*}[h]\footnotesize\centering
    \setlength{\tabcolsep}{2pt}
    \renewcommand{\arraystretch}{1.1}
    \begin{tabular}{|c|cccc|cccc|cccc|cccc|c|}
        \cline{2-18}
        \multicolumn{1}{c|}{} & \multicolumn{4}{c|}{\textbf{B}} & \multicolumn{4}{c|}{\textbf{B4}} & \multicolumn{4}{c|}{\textbf{B7}} & \multicolumn{4}{c|}{\textbf{H}} & \multirow{2}{*}{\textbf{U}}\\ 
        \cline{2-17}
        \multicolumn{1}{c|}{} & UAS & LAS & UM & LM & UAS & LAS & UM & LM& UAS & LAS & UM & LM& UAS & LAS & UM & LM & \\
        \hline 
        {\scriptsize $\rightarrow$} & 51.77 & 41.07 & 0.89 & 0.22 & 50.41 & 40.19 & 1.56 & 0.22 & 51.37 & 39.72 & 0.67 & 0.0 & 41.31 & 31.78 & 1.11 & 0.0 & 77.75 \\
        {\scriptsize $\overset{\text{\xlstm}}{\rightarrow}$} & 56.46 & 43.54 & 1.78 & 0.0 & 53.97 & 42.12 & 0.45 & 0.0 & 52.1 & 41.45 & 0.89 & 0.22 & 44.65 & 35.79 & 1.78 & 0.0 & 74.88 \\
        {\scriptsize $\leftrightarrow$} & 25.09 & 3.92 & 0.0 & 0.0 & 61.85 & 43.19 & 3.79 & 0.67 & 34.82 & 10.31 & 0.89 & 0.0 & 33.06 & 5.54 & 0.0 & 0.0 & 75.71 \\
        {\scriptsize $\overset{\text{\xlstm}}{\leftrightarrow}$} & 67.64 & 54.46 & 4.68 & 1.34 & 66.38 & 55.05 & 4.9 & 0.89 & 65.63 & 52.85 & 3.79 & 0.89 & 64.48 & 52.56 & 6.24 & 1.11 & 67.06 \\
        {\scriptsize DiT} & 60.87 & 54.76 & 1.34 & 0.22 & 79.27 & 72.16 & 15.14 & 6.46 & 71.63 & 64.94 & 8.91 & 4.9 & 28.96 & 21.62 & 1.34 & 0.45 & 95.08 \\
        {\scriptsize GaT} & \underline{\textbf{83.44}} & \underline{\textbf{77.67}} & \underline{\textbf{16.26}} & \textbf{8.02} & \textbf{85.26} & \textbf{79.51} & \underline{\textbf{24.5}} & \underline{\textbf{12.25}} & \textbf{85.19} & \textbf{79.29} & \textbf{23.16} & \textbf{13.14} & \textbf{83.59} & \textbf{77.58} & \textbf{23.16} & \textbf{11.14} & \textbf{95.81} \\
\hline 
    \end{tabular}
    \caption{\label{tab:hungarian-szeged}Performance in the Hungarian-Szeged. Same notation as in Table \ref{tab:ptb-dep}.}
\end{table*}

\begin{table*}[h]\footnotesize\centering
    \setlength{\tabcolsep}{2pt}
    \renewcommand{\arraystretch}{1.1}
    \begin{tabular}{|c|cccc|cccc|cccc|cccc|cc|}
        \cline{2-19}
        \multicolumn{1}{c|}{} & \multicolumn{4}{c|}{\textbf{B}} & \multicolumn{4}{c|}{\textbf{B4}} & \multicolumn{4}{c|}{\textbf{B7}} & \multicolumn{4}{c|}{\textbf{H}} & \multirow{2}{*}{U} & \multirow{2}{*}{\textbf{X}}\\ 
        \cline{2-17}
        \multicolumn{1}{c|}{} & UAS & LAS & UM & LM & UAS & LAS & UM & LM& UAS & LAS & UM & LM& UAS & LAS & UM & LM & &\\
        \hline 
        {\scriptsize $\rightarrow$} & 63.35 & 52.31 & 6.73 & 2.58 & 64.72 & 53.05 & 5.86 & 1.88 & 63.88 & 52.85 & 6.34 & 2.27 & 65.59 & 52.5 & 11.19 & 3.1 & 79.89 & 66.57 \\
        {\scriptsize $\overset{\text{\xlstm}}{\rightarrow}$} & 67.38 & 55.04 & 9.58 & 2.93 & 65.67 & 54.25 & 6.52 & 2.19 & 65.64 & 54.88 & 6.34 & 3.02 & 63.05 & 50.93 & 9.14 & 2.93 & 78.64 & 65.13 \\
        {\scriptsize $\leftrightarrow$} & 76.22 & 67.21 & 15.17 & 7.26 & 75.89 & 66.83 & 16.18 & 7.91 & 75.21 & 66.1 & 15.7 & 7.3 & 75.95 & 67.71 & 17.88 & 9.93 & 80.33 & 64.92 \\
        {\scriptsize $\overset{\text{\xlstm}}{\leftrightarrow}$} & 80.19 & 72.05 & 21.86 & 11.28 & 80.22 & 72.85 & 21.73 & 12.33 & 79.3 & 71.67 & 21.16 & 12.33 & 78.52 & 71.01 & 20.64 & 12.2 & 80.54 & 67.1 \\
        {\scriptsize DiT} & 84.68 & 79.75 & 31.18 & 20.33 & 85.6 & 79.67 & 33.19 & 17.93 & 85.71 & 80.87 & 34.59 & 22.52 & 87.25 & 81.52 & 38.39 & 22.96 & 91.14 & 81.4 \\
        {\scriptsize GaT} & \underline{\textbf{88.65}} & \underline{\textbf{84.28}} & \underline{\textbf{42.15}} & \underline{\textbf{29.12}} & \underline{\textbf{87.77}} & \underline{\textbf{83.43}} & \underline{\textbf{40.97}} & \underline{\textbf{28.25}} & \underline{\textbf{88.04}} & \underline{\textbf{83.79}} & \textbf{41.1} & \underline{\textbf{28.82}} & \underline{\textbf{88.09}} & \underline{\textbf{83.59}} & \underline{\textbf{42.15}} & \underline{\textbf{29.21}} & \underline{\textbf{94.67}} & \textbf{85.33} \\
\hline 
    \end{tabular}
    \caption{\label{tab:korean-kaist}Performance in the Korean-KAIST. Same notation as in Table \ref{tab:ptb-dep}.}
\end{table*}
\begin{table*}[h]\footnotesize\centering
    \setlength{\tabcolsep}{2pt}
    \renewcommand{\arraystretch}{1.1}
    \begin{tabular}{|c|cccc|cccc|cccc|cccc|cc|}
        \cline{2-19}
        \multicolumn{1}{c|}{} & \multicolumn{4}{c|}{\textbf{B}} & \multicolumn{4}{c|}{\textbf{B4}} & \multicolumn{4}{c|}{\textbf{B7}} & \multicolumn{4}{c|}{\textbf{H}} & \multirow{2}{*}{U} & \multirow{2}{*}{\textbf{X}}\\ 
        \cline{2-17}
        \multicolumn{1}{c|}{} & UAS & LAS & UM & LM & UAS & LAS & UM & LM& UAS & LAS & UM & LM& UAS & LAS & UM & LM & &\\
        \hline 
        {\scriptsize $\rightarrow$} & 68.26 & 59.84 & 10.56 & 5.91 & 67.33 & 59.39 & 9.26 & 5.64 & 66.77 & 58.62 & 8.85 & 5.1 & 56.17 & 50.04 & 8.17 & 4.7 & 90.35 & 76.23 \\
        {\scriptsize $\overset{\text{\xlstm}}{\rightarrow}$} & 72.4 & 64.15 & 11.38 & 7.54 & 69.72 & 61.34 & 10.11 & 6.0 & 69.53 & 61.41 & 9.89 & 6.23 & 60.88 & 54.99 & 11.11 & 7.27 & 89.33 & 75.44 \\
        {\scriptsize $\leftrightarrow$} & 83.69 & 73.52 & 29.71 & 14.99 & 82.71 & 72.25 & 30.11 & 14.76 & 82.63 & 71.81 & 30.43 & 14.81 & 82.22 & 72.48 & 31.06 & 15.62 & 89.32 & 77.06 \\
        {\scriptsize $\overset{\text{\xlstm}}{\leftrightarrow}$} & 88.73 & 80.4 & 41.17 & 21.99 & 89.12 & 81.04 & 42.57 & 23.43 & 88.97 & 80.64 & 42.08 & 22.26 & 88.51 & 80.3 & 43.21 & 22.71 & 91.09 & 79.47 \\
        {\scriptsize DiT} & 90.2 & 85.45 & 50.74 & 32.37 & 94.51 & 89.06 & 63.61 & 35.67 & 94.42 & 89.3 & 62.8 & 37.43 & 94.36 & 87.56 & 62.75 & 30.29 & 97.53 & 92.58 \\
        {\scriptsize GaT} & \underline{\textbf{95.4}} & \underline{\textbf{91.01}} & \underline{\textbf{64.97}} & \underline{\textbf{40.77}} & \underline{\textbf{94.78}} & \underline{\textbf{90.38}} & \textbf{64.6} & \underline{\textbf{40.5}} & \textbf{95.39} & \underline{\textbf{90.96}} & \underline{\textbf{66.32}} & \underline{\textbf{41.4}} & \textbf{95.07} & \underline{\textbf{90.69}} & \underline{\textbf{66.86}} & \underline{\textbf{41.31}} & \textbf{98.82} & \underline{\textbf{96.13}} \\
\hline 
    \end{tabular}
    \caption{\label{tab:polish-pdb}Performance in the Polish-PDB. Same notation as in Table \ref{tab:ptb-dep}.}
\end{table*}
\begin{table*}[h]\footnotesize\centering
    \setlength{\tabcolsep}{2pt}
    \renewcommand{\arraystretch}{1.1}
    \begin{tabular}{|c|cccc|cccc|cccc|cccc|cc|}
        \cline{2-19}
        \multicolumn{1}{c|}{} & \multicolumn{4}{c|}{\textbf{B}} & \multicolumn{4}{c|}{\textbf{B4}} & \multicolumn{4}{c|}{\textbf{B7}} & \multicolumn{4}{c|}{\textbf{H}} & \multirow{2}{*}{U} & \multirow{2}{*}{\textbf{X}}\\ 
        \cline{2-17}
        \multicolumn{1}{c|}{} & UAS & LAS & UM & LM & UAS & LAS & UM & LM& UAS & LAS & UM & LM& UAS & LAS & UM & LM & &\\
        \hline 
        {\scriptsize $\rightarrow$} & 64.75 & 54.52 & 9.52 & 2.63 & 67.85 & 57.87 & 11.89 & 4.18 & 66.94 & 56.96 & 11.24 & 4.1 & 59.21 & 52.49 & 8.29 & 4.27 & 88.74 & 83.39 \\
        {\scriptsize $\overset{\text{\xlstm}}{\rightarrow}$} & 69.26 & 59.46 & 9.93 & 3.45 & 69.88 & 60.32 & 10.58 & 3.69 & 69.99 & 59.95 & 10.66 & 3.61 & 62.8 & 56.48 & 9.35 & 4.1 & 84.87 & 79.29 \\
        {\scriptsize $\leftrightarrow$} & 74.69 & 65.8 & 13.62 & 7.22 & 77.38 & 69.74 & 17.64 & 10.5 & 76.44 & 67.34 & 16.98 & 8.94 & 72.93 & 64.06 & 19.52 & 7.88 & 84.5 & 80.17 \\
        {\scriptsize $\overset{\text{\xlstm}}{\leftrightarrow}$} & 82.93 & 76.48 & 22.72 & 14.27 & 84.38 & 78.06 & 27.97 & 17.06 & 84.09 & 77.3 & 25.76 & 16.0 & 82.38 & 76.12 & 28.71 & 16.9 & 90.8 & 82.32 \\
        {\scriptsize DiT} & 74.22 & 72.02 & 17.15 & 14.68 & 92.47 & 89.54 & 53.24 & 41.18 & 91.3 & 88.52 & 49.38 & 38.23 & 90.49 & 84.5 & 44.63 & 23.46 & 97.67 & 92.14 \\
        {\scriptsize GaT} & \textbf{93.08} & \textbf{90.55} & \underline{\textbf{53.24}} & \underline{\textbf{42.74}} & \textbf{93.34} & \textbf{90.87} & \textbf{56.77} & \textbf{45.04} & \underline{\textbf{93.91}} & \underline{\textbf{91.6}} & \underline{\textbf{58.57}} & \underline{\textbf{47.83}} & \underline{\textbf{93.2}} & \underline{\textbf{90.42}} & \underline{\textbf{56.19}} & \underline{\textbf{43.4}} & \textbf{98.18} & \underline{\textbf{95.89}} \\
\hline 
    \end{tabular}
    \caption{\label{tab:swedish-talbanken}Performance in the Swedish-Talbanken. Same notation as in Table \ref{tab:ptb-dep}.}
\end{table*}

\begin{table*}[h]\footnotesize\centering
    \setlength{\tabcolsep}{2pt}
    \renewcommand{\arraystretch}{1.1}
    \begin{tabular}{|c|cccc|cccc|cccc|cccc|}
        \cline{2-17}
        \multicolumn{1}{c|}{} & \multicolumn{4}{c|}{\textbf{R}} & \multicolumn{4}{c|}{\textbf{B}} & \multicolumn{4}{c|}{$\boldsymbol{4k}$} & \multicolumn{4}{c|}{$\boldsymbol{6k}$} \\
        \cline{2-17}
        \multicolumn{1}{c|}{} & UF & LF & UM & LM & UF & LF & UM & LM& UF & LF & UF & LF & UF & LF & UM & LM \\
        \hline 
        {\scriptsize $\rightarrow$} & 67.2 & 59.67 & 0.99 & 0.92 & 64.91 & 55.5 & 3.4 & 2.41 & 56.53 & 43.34 & 1.35 & 0.92 & 63.11 & 54.79 & 3.26 & 2.2 \\
        {\scriptsize $\overset{\text{\xlstm}}{\rightarrow}$} & 70.81 & 65.3 & 1.56 & 1.13 & 67.55 & 60.52 & 3.97 & 2.98 & 59.01 & 45.98 & 1.28 & 1.06 & 66.25 & 59.24 & 3.48 & 2.55 \\
        {\scriptsize $\leftrightarrow$} & 85.81 & 80.5 & 17.3 & 13.69 & 87.26 & 82.81 & 24.04 & 19.65 & 86.38 & 82.2 & 20.0 & 17.09 & 87.07 & 83.02 & 22.48 & 18.72 \\
        {\scriptsize $\overset{\text{\xlstm}}{\leftrightarrow}$} & 88.57 & 84.53 & 23.62 & 19.22 & 90.31 & 86.7 & 33.19 & 26.88 & 89.38 & 85.77 & 27.16 & 23.19 & 89.36 & 85.58 & 29.36 & 23.26 \\
        {\scriptsize DiT} & 79.9 & 75.05 & 16.81 & 14.54 & 93.16 & 85.91 & 50.28 & 36.03 & 90.84 & 84.73 & 37.87 & 31.84 & 92.6 & 85.73 & 48.65 & 36.38 \\
        {\scriptsize GaT} & \textbf{91.74} & \underline{\textbf{89.77}} & \underline{\textbf{35.53}} & \underline{\textbf{32.2}} & \textbf{95.31} & \underline{\textbf{93.57}} & \underline{\textbf{56.6}} & \underline{\textbf{49.57}} & \underline{\textbf{94.57}} & \underline{\textbf{92.82}} & \underline{\textbf{48.94}} & \underline{\textbf{43.83}} & \textbf{95.07} & \underline{\textbf{93.42}} & \underline{\textbf{55.82}} & \underline{\textbf{49.5}} \\
\hline 
    \end{tabular}
    \caption{\label{tab:en-dm}Performance in the English-DM dataset. Same abbreviations as in Table \ref{tab:graph-results}: relative (\textbf{R}), bracketing (\textbf{B}), $4k$-bit  ($\boldsymbol{4k}$) and $6k$-bit ($\boldsymbol{6k}$) encodings.}
\end{table*}

\begin{table*}[h]\footnotesize\centering
    \setlength{\tabcolsep}{2pt}
    \renewcommand{\arraystretch}{1.1}
    \begin{tabular}{|c|cccc|cccc|cccc|cccc|}
        \cline{2-17}
        \multicolumn{1}{c|}{} & \multicolumn{4}{c|}{\textbf{R}} & \multicolumn{4}{c|}{\textbf{B}} & \multicolumn{4}{c|}{$\boldsymbol{4k}$} & \multicolumn{4}{c|}{$\boldsymbol{6k}$} \\
        \cline{2-17}
        \multicolumn{1}{c|}{} & UF & LF & UM & LM & UF & LF & UM & LM& UF & LF & UF & LF & UF & LF & UM & LM \\
        \hline 
        {\scriptsize $\rightarrow$} & 71.31 & 63.64 & 4.56 & 3.53 & 67.05 & 60.02 & 4.26 & 3.24 & 70.43 & 60.64 & 3.53 & 2.35 & 68.89 & 61.7 & 4.12 & 2.94 \\
        {\scriptsize $\overset{\text{\xlstm}}{\rightarrow}$} & 75.02 & 67.86 & 5.59 & 3.53 & 69.9 & 63.17 & 5.29 & 4.85 & 71.57 & 63.14 & 5.29 & 4.26 & 69.45 & 62.62 & 4.85 & 3.68 \\
        {\scriptsize $\leftrightarrow$} & 76.92 & 68.14 & 7.21 & 5.44 & 77.33 & 68.51 & 7.79 & 6.03 & 75.95 & 67.45 & 7.79 & 5.88 & 75.76 & 67.07 & 7.79 & 5.44 \\
        {\scriptsize $\overset{\text{\xlstm}}{\leftrightarrow}$} & 80.24 & 72.53 & 9.41 & 6.91 & 80.77 & 73.15 & 10.74 & 6.47 & 80.49 & 72.85 & 10.15 & 6.76 & 80.35 & 73.31 & 8.97 & 6.47 \\
        {\scriptsize DiT} & 78.79 & 72.12 & 8.09 & 5.0 & 77.88 & 69.52 & 8.97 & 5.59 & 82.91 & 73.37 & 9.85 & 5.44 & 85.31 & 76.61 & 15.0 & 7.79 \\
        {\scriptsize GaT} & \textbf{84.85} & \underline{\textbf{78.24}} & \underline{\textbf{11.91}} & \underline{\textbf{8.53}} & \textbf{87.66} & \textbf{80.39} & \textbf{17.5} & \textbf{8.68} & \underline{\textbf{88.07}} & \underline{\textbf{81.3}} & \textbf{17.79} & \underline{\textbf{10.29}} & \underline{\textbf{88.52}} & \underline{\textbf{81.75}} & \underline{\textbf{18.97}} & \underline{\textbf{10.29}} \\
\hline 
    \end{tabular}
    \caption{\label{tab:iwpt-2021-ar-padt}Graph parsing performance in the Arabic-PADT (IWPT) dataset. Same notation as in Table \ref{tab:en-dm}.}
\end{table*}

\begin{table*}[h]\footnotesize\centering
    \setlength{\tabcolsep}{2pt}
    \renewcommand{\arraystretch}{1.1}
    \begin{tabular}{|c|cccc|cccc|cccc|cccc|}
        \cline{2-17}
        \multicolumn{1}{c|}{} & \multicolumn{4}{c|}{\textbf{R}} & \multicolumn{4}{c|}{\textbf{B}} & \multicolumn{4}{c|}{$\boldsymbol{4k}$} & \multicolumn{4}{c|}{$\boldsymbol{6k}$} \\
        \cline{2-17}
        \multicolumn{1}{c|}{} & UF & LF & UM & LM & UF & LF & UM & LM& UF & LF & UF & LF & UF & LF & UM & LM \\
        \hline 
        {\scriptsize $\rightarrow$} & 58.36 & 50.5 & 4.03 & 2.06 & 68.51 & 60.46 & 7.35 & 3.67 & 70.72 & 62.28 & 8.51 & 4.66 & 70.38 & 61.84 & 7.53 & 4.12 \\
        {\scriptsize $\overset{\text{\xlstm}}{\rightarrow}$} & 65.33 & 60.0 & 5.91 & 4.57 & 74.06 & 67.5 & 11.47 & 7.62 & 74.85 & 67.2 & 9.77 & 5.82 & 74.54 & 67.78 & 11.83 & 7.17 \\
        {\scriptsize $\leftrightarrow$} & 83.5 & 73.55 & 25.45 & 11.47 & 81.41 & 72.37 & 21.33 & 9.77 & 78.92 & 69.29 & 19.98 & 9.77 & 82.24 & 72.94 & 25.27 & 11.74 \\
        {\scriptsize $\overset{\text{\xlstm}}{\leftrightarrow}$} & 88.06 & 81.94 & 33.96 & 20.97 & 88.21 & 81.65 & 34.5 & 20.79 & 88.51 & 82.6 & 38.26 & 23.48 & 88.99 & 82.78 & 38.71 & 22.85 \\
        {\scriptsize DiT} & 84.75 & 80.65 & 25.72 & 17.56 & 86.15 & 81.2 & 35.22 & 23.57 & 91.37 & 85.56 & 47.58 & 28.41 & 92.79 & 86.83 & 56.99 & 32.26 \\
        {\scriptsize GaT} & \underline{\textbf{91.84}} & \underline{\textbf{87.81}} & \underline{\textbf{47.31}} & \underline{\textbf{32.17}} & \underline{\textbf{94.46}} & \underline{\textbf{90.34}} & \underline{\textbf{59.32}} & \underline{\textbf{39.78}} & \textbf{94.86} & \textbf{90.47} & \textbf{62.46} & \textbf{39.52} & \underline{\textbf{95.04}} & \textbf{90.43} & \textbf{64.34} & \underline{\textbf{40.86}} \\
\hline 
    \end{tabular}
    \caption{\label{tab:iwpt-2021-bg-btb}Graph parsing performance in the Bulgarian-BTB (IWPT) dataset. Same notation as in Table \ref{tab:en-dm}.}
\end{table*}

\begin{table*}[h]\footnotesize\centering
    \setlength{\tabcolsep}{2pt}
    \renewcommand{\arraystretch}{1.1}
    \begin{tabular}{|c|cccc|cccc|cccc|cccc|}
        \cline{2-17}
        \multicolumn{1}{c|}{} & \multicolumn{4}{c|}{\textbf{R}} & \multicolumn{4}{c|}{\textbf{B}} & \multicolumn{4}{c|}{$\boldsymbol{4k}$} & \multicolumn{4}{c|}{$\boldsymbol{6k}$} \\
        \cline{2-17}
        \multicolumn{1}{c|}{} & UF & LF & UM & LM & UF & LF & UM & LM& UF & LF & UF & LF & UF & LF & UM & LM \\
        \hline 
        {\scriptsize $\rightarrow$} & 67.16 & 60.9 & 14.25 & 11.84 & 73.39 & 66.31 & 14.69 & 13.16 & 77.13 & 69.3 & 15.79 & 13.82 & 76.75 & 69.13 & 14.69 & 12.5 \\
        {\scriptsize $\overset{\text{\xlstm}}{\rightarrow}$} & 69.89 & 64.98 & 13.82 & 12.5 & 77.03 & 70.91 & 16.67 & 14.91 & 78.19 & 70.58 & 16.01 & 14.69 & 77.62 & 70.95 & 16.67 & 14.47 \\
        {\scriptsize $\leftrightarrow$} & 79.88 & 72.09 & 19.96 & 16.67 & 80.99 & 74.33 & 21.49 & 18.42 & 81.71 & 74.19 & 24.56 & 18.86 & 0.26 & 0.0 & 0.0 & 0.0 \\
        {\scriptsize $\overset{\text{\xlstm}}{\leftrightarrow}$} & 84.91 & 80.24 & 23.46 & 21.05 & 84.37 & 79.88 & 21.49 & 18.42 & 85.95 & 81.25 & 26.32 & 23.46 & 86.66 & 82.21 & 28.07 & 24.56 \\
        {\scriptsize DiT} & 77.62 & 75.55 & 16.01 & 15.13 & 69.1 & 65.97 & 13.38 & 12.72 & 86.52 & 81.72 & 34.87 & 25.22 & 87.35 & 83.29 & 37.06 & 28.29 \\
        {\scriptsize GaT} & \textbf{86.81} & \textbf{84.49} & \textbf{26.75} & \textbf{23.9} & \underline{\textbf{93.16}} & \underline{\textbf{90.39}} & \underline{\textbf{42.76}} & \underline{\textbf{35.31}} & \underline{\textbf{93.65}} & \underline{\textbf{90.84}} & \underline{\textbf{49.78}} & \underline{\textbf{40.13}} & \underline{\textbf{93.92}} & \underline{\textbf{91.73}} & \underline{\textbf{49.78}} & \underline{\textbf{42.32}} \\
\hline 
    \end{tabular}
    \caption{\label{tab:iwpt-2021-fr-sequoia}Graph parsing performance in the French-Sequoia (IWPT) dataset. Same notation as in Table \ref{tab:en-dm}.}
\end{table*}

\begin{table*}[h]\footnotesize\centering
    \setlength{\tabcolsep}{2pt}
    \renewcommand{\arraystretch}{1.1}
    \begin{tabular}{|c|cccc|cccc|cccc|cccc|}
        \cline{2-17}
        \multicolumn{1}{c|}{} & \multicolumn{4}{c|}{\textbf{R}} & \multicolumn{4}{c|}{\textbf{B}} & \multicolumn{4}{c|}{$\boldsymbol{4k}$} & \multicolumn{4}{c|}{$\boldsymbol{6k}$} \\
        \cline{2-17}
        \multicolumn{1}{c|}{} & UF & LF & UM & LM & UF & LF & UM & LM& UF & LF & UF & LF & UF & LF & UM & LM \\
        \hline 
        {\scriptsize $\rightarrow$} & 66.75 & 60.39 & 5.19 & 3.11 & 75.99 & 68.58 & 9.13 & 5.19 & 76.82 & 68.23 & 5.6 & 3.32 & 76.31 & 69.11 & 9.54 & 6.43 \\
        {\scriptsize $\overset{\text{\xlstm}}{\rightarrow}$} & 70.99 & 66.04 & 8.51 & 4.98 & 77.53 & 71.37 & 10.79 & 6.64 & 78.87 & 71.48 & 11.2 & 8.71 & 78.84 & 72.83 & 12.24 & 7.88 \\
        {\scriptsize $\leftrightarrow$} & 87.33 & 81.65 & 29.25 & 17.84 & 86.9 & 81.49 & 29.46 & 19.92 & 85.9 & 79.92 & 28.42 & 18.46 & 86.28 & 80.08 & 29.46 & 16.8 \\
        {\scriptsize $\overset{\text{\xlstm}}{\leftrightarrow}$} & 89.61 & 85.53 & 35.48 & 26.76 & 90.18 & 86.16 & 37.34 & 28.63 & 90.22 & 86.27 & 37.97 & 29.25 & 90.59 & 86.69 & 39.83 & 29.88 \\
        {\scriptsize DiT} & 80.72 & 77.7 & 9.34 & 6.02 & 85.76 & 82.61 & 31.95 & 25.31 & 91.11 & 86.7 & 41.29 & 29.88 & 92.69 & 87.85 & 48.96 & 31.33 \\
        {\scriptsize GaT} & \underline{\textbf{91.99}} & \underline{\textbf{89.52}} & \underline{\textbf{43.15}} & \underline{\textbf{34.23}} & \textbf{94.0} & \underline{\textbf{91.23}} & \textbf{50.41} & \underline{\textbf{39.42}} & \textbf{94.17} & \textbf{91.46} & \underline{\textbf{56.43}} & \underline{\textbf{43.15}} & \textbf{94.19} & \textbf{91.33} & \textbf{54.98} & \underline{\textbf{42.53}} \\
\hline 
    \end{tabular}
    \caption{\label{tab:iwpt-2021-it-isdt}Graph parsing performance in the Italian-ISDT (IWPT) dataset. Same notation as in Table \ref{tab:en-dm}.}
\end{table*}

\begin{table*}[h]\footnotesize\centering
    \setlength{\tabcolsep}{2pt}
    \renewcommand{\arraystretch}{1.1}
    \begin{tabular}{|c|cccc|cccc|cccc|cccc|}
        \cline{2-17}
        \multicolumn{1}{c|}{} & \multicolumn{4}{c|}{\textbf{R}} & \multicolumn{4}{c|}{\textbf{B}} & \multicolumn{4}{c|}{$\boldsymbol{4k}$} & \multicolumn{4}{c|}{$\boldsymbol{6k}$} \\
        \cline{2-17}
        \multicolumn{1}{c|}{} & UF & LF & UM & LM & UF & LF & UM & LM& UF & LF & UF & LF & UF & LF & UM & LM \\
        \hline 
        {\scriptsize $\rightarrow$} & 49.11 & 40.97 & 1.17 & 0.17 & 59.27 & 50.18 & 3.02 & 0.67 & 61.99 & 50.9 & 2.85 & 0.67 & 63.1 & 53.47 & 2.35 & 0.84 \\
        {\scriptsize $\overset{\text{\xlstm}}{\rightarrow}$} & 54.77 & 47.67 & 1.17 & 0.67 & 65.79 & 58.64 & 3.52 & 1.68 & 66.01 & 57.26 & 2.18 & 0.67 & 66.56 & 59.3 & 4.03 & 1.85 \\
        {\scriptsize $\leftrightarrow$} & 79.69 & 69.01 & 15.94 & 6.71 & 2.43 & 0.0 & 0.0 & 0.0 & 71.69 & 62.38 & 11.58 & 5.54 & 74.03 & 63.54 & 11.41 & 5.2 \\
        {\scriptsize $\overset{\text{\xlstm}}{\leftrightarrow}$} & 83.47 & 77.19 & 20.97 & 13.59 & 83.29 & 76.5 & 20.13 & 13.26 & 82.29 & 75.51 & 23.49 & 13.59 & 83.31 & 76.64 & 22.65 & 14.6 \\
        {\scriptsize DiT} & 76.72 & 72.07 & 8.22 & 5.03 & 75.04 & 69.88 & 13.42 & 7.89 & 85.06 & 78.46 & 31.71 & 19.46 & 88.34 & 81.7 & 44.46 & 22.82 \\
        {\scriptsize GaT} & \underline{\textbf{88.52}} & \underline{\textbf{84.1}} & \underline{\textbf{32.89}} & \underline{\textbf{22.15}} & \textbf{92.11} & \underline{\textbf{87.14}} & \underline{\textbf{48.32}} & \underline{\textbf{30.37}} & \textbf{91.56} & \textbf{87.03} & \underline{\textbf{52.52}} & \underline{\textbf{33.05}} & \textbf{92.18} & \textbf{87.75} & \textbf{52.52} & \underline{\textbf{34.9}} \\
\hline 
    \end{tabular}
    \caption{\label{tab:iwpt-2021-nl-alpino}Graph parsing performance in the Dutch-Alpino (IWPT) dataset. Same notation as in Table \ref{tab:en-dm}.}
\end{table*}

\begin{table*}[h]\footnotesize\centering
    \setlength{\tabcolsep}{2pt}
    \renewcommand{\arraystretch}{1.1}
    \begin{tabular}{|c|cccc|cccc|cccc|cccc|}
        \cline{2-17}
        \multicolumn{1}{c|}{} & \multicolumn{4}{c|}{\textbf{R}} & \multicolumn{4}{c|}{\textbf{B}} & \multicolumn{4}{c|}{$\boldsymbol{4k}$} & \multicolumn{4}{c|}{$\boldsymbol{6k}$} \\
        \cline{2-17}
        \multicolumn{1}{c|}{} & UF & LF & UM & LM & UF & LF & UM & LM& UF & LF & UF & LF & UF & LF & UM & LM \\
        \hline 
        {\scriptsize $\rightarrow$} & 42.26 & 14.6 & 0.0 & 0.0 & 28.78 & 5.54 & 0.0 & 0.0 & 29.3 & 4.39 & 0.0 & 0.0 & 38.1 & 6.88 & 0.0 & 0.0 \\
        {\scriptsize $\overset{\text{\xlstm}}{\rightarrow}$} & 55.25 & 40.02 & 0.0 & 0.0 & 51.9 & 34.11 & 0.0 & 0.0 & 58.18 & 37.85 & 0.83 & 0.0 & 57.91 & 40.49 & 0.83 & 0.0 \\
        {\scriptsize $\leftrightarrow$} & 38.64 & 8.53 & 0.0 & 0.0 & 7.37 & 7.37 & 0.0 & 0.0 & 35.17 & 6.59 & 0.0 & 0.0 & 36.06 & 1.75 & 0.0 & 0.0 \\
        {\scriptsize $\overset{\text{\xlstm}}{\leftrightarrow}$} & 63.23 & 48.91 & \textbf{0.83} & \underline{\textbf{0.83}} & 65.48 & 47.64 & 2.5 & 0.0 & 68.24 & 52.21 & 3.33 & \textbf{0.83} & 67.0 & 51.61 & 1.67 & 0.83 \\
        {\scriptsize DiT} & 25.94 & 2.95 & 0.0 & 0.0 & 42.2 & 12.44 & 0.0 & 0.0 & 20.96 & 3.97 & 0.0 & 0.0 & 19.4 & 3.11 & 0.0 & 0.0 \\
        {\scriptsize GaT} & \underline{\textbf{66.45}} & \underline{\textbf{55.61}} & 0.83 & 0.0 & \textbf{74.03} & \textbf{60.52} & \underline{\textbf{6.67}} & \underline{\textbf{0.83}} & \textbf{76.54} & \textbf{62.32} & \textbf{8.33} & 0.83 & \textbf{76.29} & \textbf{62.79} & \underline{\textbf{12.5}} & \underline{\textbf{2.5}} \\
\hline 
    \end{tabular}
    \caption{\label{tab:iwpt-2021-ta-ttb}Graph parsing performance in the Tamil-TTB (IWPT) dataset. Same notation as in Table \ref{tab:en-dm}.}
\end{table*}
\end{document}